\documentclass[11pt]{article}

\usepackage[preprint]{acl}

\usepackage{times}
\usepackage{latexsym}

\usepackage[T1]{fontenc}

\usepackage[utf8]{inputenc}

\usepackage{microtype}

\usepackage{inconsolata}

\usepackage{graphicx}
\usepackage{subcaption}

\usepackage[T1]{fontenc}    
\usepackage{url}            
\usepackage{booktabs}       
\usepackage{amsfonts}       
\usepackage{nicefrac}       
\usepackage{microtype}      
\usepackage{xcolor}         
\usepackage{xspace}

\usepackage{multirow}
\usepackage{xcolor}
\usepackage{colortbl}
\usepackage{graphicx}
\usepackage{wrapfig}
\usepackage{amsmath}
\usepackage{listings}
\usepackage{bm}

\usepackage{paralist}
\usepackage{algorithm}
\usepackage{algorithmic}
\usepackage{amsmath}
\usepackage{multirow}
\usepackage{booktabs}
\usepackage{colortbl}

\usepackage{float}
\usepackage{graphicx,overpic}
\usepackage{amsmath}
\usepackage{colortbl}
\usepackage{color,xcolor}
\definecolor{light-gray}{gray}{0.95}    
\usepackage{multirow}
\usepackage{bm}
\usepackage{enumerate}
\usepackage{enumitem}
\usepackage{pifont}
\usepackage{amsthm} 
\usepackage{makecell}
\usepackage{pifont}

\theoremstyle{definition}

\usepackage{tikz}
\usepackage[framemethod=default]{mdframed}

\usepackage{framed}
\usepackage{mathtools}

\usepackage[most]{tcolorbox}

\definecolor{cvprblue}{rgb}{0.21,0.49,0.74}
\definecolor{lightgreen3}{RGB}{236,245,228}
\definecolor{lightyellow}{RGB}{255,255,210} 
\definecolor{lightorange}{RGB}{255,235,210} 
\definecolor{default}{RGB}{255,255,255}
\definecolor{pseudoblue}{RGB}{55,113,175}
\definecolor{mgreen}{RGB}{6,128,67}
\definecolor{mgray}{RGB}{128,128,128}
\definecolor{mygreen}{RGB}{233,247,234}
\definecolor{mygray}{gray}{0.97}
\definecolor{mgray}{RGB}{240,240,240}
\definecolor{purple}{RGB}{128,128,247}
\definecolor{mypink}{RGB}{252,237,236}
\definecolor{pink2}{RGB}{234,223,222}
\definecolor{pink3}{RGB}{237,227,228}
\definecolor{blue1}{RGB}{229,243,247}

\NewDocumentCommand{\circnumgreen}{O{teal!18} m}{%
  \tikz[baseline=(char.base)]{
    \node[shape=circle, fill=#1, inner sep=1pt, font=\small\bfseries] (char)
      {\textcolor{teal!70!black}{#2}};
  }%
}

\NewDocumentCommand{\circnumpink}{O{pink!30!white} m}{%
  \tikz[baseline=(char.base)]{
    \node[shape=circle, fill=#1, inner sep=1pt, font=\small\bfseries] (char)
      {\textcolor{pink!70!red}{#2}};
  }%
}

\definecolor{myblue}{RGB}{41, 41, 148} 

\newtcolorbox{redquestion}[1][]{
  enhanced,
  boxrule=0pt, 
  leftrule=4pt,           
  frame hidden,
  colback=mypink,           
  colframe=pink,      
  sharp corners,
  left=12pt, right=12pt, top=8pt, bottom=8pt,
  parbox=false,
  before upper={\textbf{\textcolor{coralred}{Question.}}~}, 
  #1
}

\newtcolorbox{blquestion}[1][]{
  enhanced,
  boxrule=0pt,          
  leftrule=4pt, 
  frame hidden,         
  colback=softblue,     
  colframe=myblue,     
  sharp corners,        
  left=12pt, right=12pt, top=8pt, bottom=8pt, 
  parbox=false,
  before upper={\textbf{\textcolor{mainred}{Question.}}~}, 
  #1
}
\title{Double: Breaking the Acceleration Limit via Double Retrieval Speculative Parallelism}

\author{
   Yuhao Shen$^{\hspace{0.4em}\mathsection}$ \quad Tianyu Liu$^{\hspace{0.4em}\diamondsuit}$ \quad Junyi Shen$^{\ddag}$
  \quad
  \textbf{Jinyang Wu}$^{\clubsuit}$
  \quad
  \\
  \textbf{Quan Kong}$^{\mathsection}$
  \quad
  \textbf{Huan Li}$^{\mathsection}$
  \quad
  \textbf{Cong Wang}\thanks{The Corresponding Authors.}$^{\hspace{0.4em}\mathsection}$
  \\
  ${}^\mathsection$Zhejiang University \quad 
  ${}^\diamondsuit$University of Science and Technology of China \\
  ${}^\ddag$National University of Singapore\quad 
  ${}^\clubsuit$Tsinghua University\\
    \texttt{\{riven, quankong, lihuan.cs, cwang85\}@zju.edu.cn} \quad \\
    \texttt{tianyu\_liu@mail.ustc.edu.cn}\quad 
    \texttt{j1shen@comp.nus.edu.sg}\\
    \texttt{wu-jy23@mails.tsinghua.edu.cn}\\
}

\begin{document}
\maketitle
\begin{abstract}
Parallel Speculative Decoding (PSD) accelerates traditional Speculative Decoding (SD) by overlapping draft generation with verification. However, it remains hampered by two fundamental challenges: (1) a theoretical speedup ceiling dictated by the speed ratio between the draft and target models, and (2) high computational waste and pipeline stall due to mid-sequence token rejections of early errors. To address these limitations, we introduce \textsc{Double} (Double Retrieval Speculative Parallelism). By bridging the gap between SD and PSD, our framework resolves the Retrieval \emph{Precision-Efficiency Dilemma} through a novel synchronous mechanism. Specifically, we enable the draft model to execute iterative retrieval speculations to break the theoretical speedup limits; to alleviate rejections without rollback, the target model performs authoritative retrieval to generate multi-token guidance. \textsc{Double} is entirely training-free and lossless. Extensive experiments demonstrate state-of-the-art speedup of $\textbf{5.3}\times$ on LLaMA3.3-70B and $\textbf{2.8}\times$ on Qwen3-32B, significantly outperforming the advanced method EAGLE-3 that requires extensive model training. Our code is available at \url{https://github.com/Sylvan820/Double1}.
\end{abstract}

\begin{figure}[!ht]
\centering
\includegraphics[width=\columnwidth]{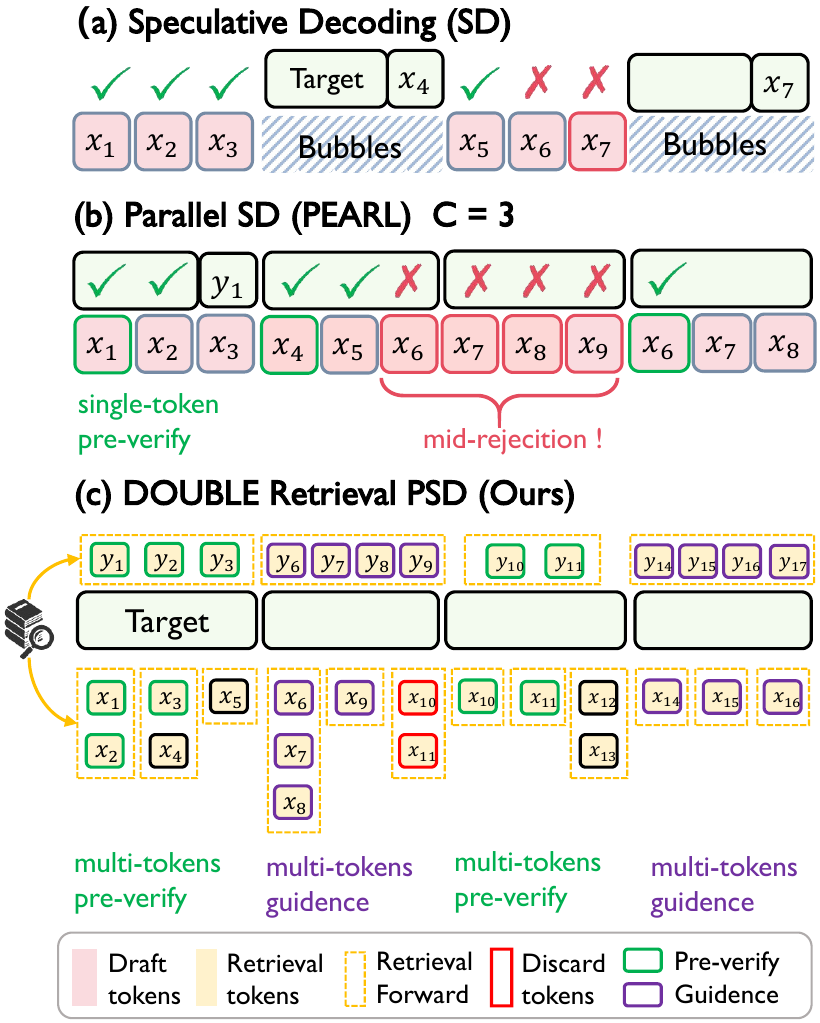} 
\caption{Comparison between SD, PSD and \textsc{Double}. 
\textbf{(a)} SD suffers from pipeline bubbles due to sequential dependency. \textbf{(b)} PSD overlaps these processes to reduce latency but struggles with \textit{mid-sequence rejection}, where tokens generated are wasted after an early error (e.g., red boxes $x_{6-9}$).
\textbf{(c)} \textsc{Double} resolves these issues through a double-retrieval mechanism: draft model leverages retrieval to expand draft length (more tokens $x_{1\text{-}5}$); target model performs retrieval to offer \textbf{multi-token pre-verify and guidance} (retrieve $y_{1\text{-}3}$ for verifying $x_{1\text{-}3}$) to ensure precision, thereby breaking the speed limit $C$ jointly and mitigating rejection penalties.
}
\label{fig:intro}
\vspace*{-0.3in}
\end{figure}


\section{Introduction}

The sequential nature of autoregressive decoding inherently restricts parallelization as each token's generation is contingent upon its predecessors~\citep{brown2020language}. While techniques like quantization, distillation, and efficient attention~\citep{hinton2015distilling,dao2022flashattention, choi2018pact} alleviate computational burdens, inference remains bottlenecked by memory bandwidth.



 \begin{figure*}[t]
\centering
\includegraphics[width=\textwidth]{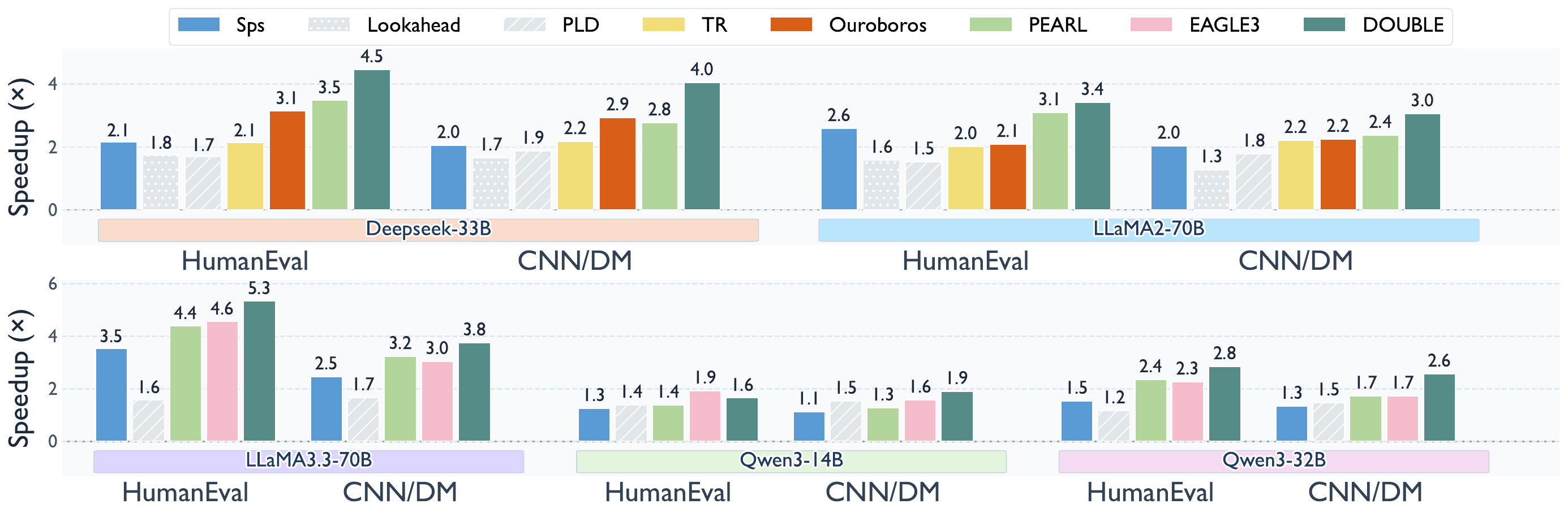} 
\vspace*{-0.2in}
\caption{Speedup ratios of different methods on HumanEval and CNN/DM. \textsc{Double} achieves a speedup of 5.3$\times$ on LLama3.3-70B and 2.8$\times$ on Qwen3-32B over than EAGLE3. Full results are available in Table \ref{tab:main_resultes}.
}
\label{fig:performance}
\vspace*{-0.2in}
\end{figure*}

Speculative Decoding (SD) has emerged as a promising lossless approach to address this memory-bound limitation~\citep{leviathan2023fast, chen2023accelerating}. SD utilizes a lightweight \emph{draft model} to propose a sequence of $\gamma$ tokens, which are then verified in parallel by the larger \emph{target model}. However, the predominant draft-then-verify paradigm in most SD frameworks introduces a significant mutual waiting bottleneck. As illustrated in Fig.~\ref{fig:intro}, this sequential dependency between the draft and verification stages leads to \emph{Bubbles} with underutilized GPU resources. To enhance hardware efficiency, Parallel Speculative Decoding (PSD) frameworks like PEARL~\citep{liu2024parallel} and SpecBranch~\citep{shen2025speculative} have demonstrated substantial speedup by adopting a draft-when-verify approach, enabling the draft and verification processes to proceed concurrently. Despite these advancements, existing PSD methods face two primary challenges:

\begin{itemize}[leftmargin=*, itemsep=0pt, topsep=4pt]
    \item PSD is strictly constrained by a \textbf{theoretical upper speedup $C$} (the draft-to-target latency ratio). Unlike SD which employs tailored distilled models~\citep{li2025eagle}, PSD utilizes off-the-shelf small models ($0.6$B, $1.3$B, $7$B). This limits $C$ to a modest $1.5-5\times$ relative to large targets ($14$B, $33$B, $70$B). Fig~\ref{fig:performance} shows PEARL achieves a marginal speedup $1.3\times$ on Qwen-0.6B/14B models ($C=1.6$) due to the autoregressive drafting.

    \item Existing frameworks struggle with \textbf{mid-sequence token rejection}. As shown in Fig.~\ref{fig:intro}, although PEARL employs additional target forwards for pre-verify, it incurs high overhead with low returns for single-token (only $y_1$ for $x_1$) predictions, causing the rejection of mid tokens (red boxes $x_6\text{-}x_9$). While SpecBranch attempts to mitigate this via an offline predictor, it introduces additional training costs and yields lossy predictions compared to the target model.
\end{itemize}

\textcolor{teal}{Can we generate longer and high-quality multi-token proposals without paying {\emph{autoregressive}} draft model compute per token, while keeping high acceptance?}

Retrieval is a promising technique~\citep{he2023rest}: by reusing previously seen continuations, we can propose multiple tokens with minimal neural computation. However, existing retrieval decoding methods face a \textbf{Retrieval Precision-Efficiency Dilemma} when combined with speculation. Using retrieval only at the target side is precise but offers limited acceleration~\citep{pld-saxena-2023,token-recycle-luo-2024,liu2025logitspec,fu2024break}, because the expensive target forward pass still remains; using retrieval at the draft side is desirable but often imprecise~\citep{zhao2024ouroboros}, increasing rejection rates and causing parallel pipeline stalls. As a result, prior approaches typically improve either proposal speed or quality, but NOT both, that remain under the PSD ceiling and rejection rollback.


Based on these insights, we introduce \textsc{Double}, which inherits the parallel pipeline but comes with a new double retrieval design on both sides of the draft/target models. Specifically, the draft model executes $\gamma$ iterations of retrieval-based speculation to replace the slow auto-regressive generation, which effectively breaks the speedup limit $C$. Meanwhile, the target model performs a single-step retrieval to generate multiple tokens, in order to guide the draft process and mitigate mid-sequence rejections. Our contributions are summarized:
\begin{itemize}[leftmargin=*, itemsep=0pt, topsep=4pt]
    \item We extend the theoretical speculation process from single-round to multi-round, which unifies both SD and PSD frameworks with a formal proof of the speed limit $C$. This has laid a theoretical foundation for retrieval.  
    \item We propose a novel Double Retrieval mechanism that resolves the \emph{Precision-Efficiency Dilemma} on parallel architecture. This strategy significantly enlarges the drafting length and leverages the power of target model for multi-token verification and forward guidance.
    \item Across multiple datasets and model pairs, \textsc{Double} achieves a SOTA speedup of \textbf{5.3}$\times$ on LLama3.3-70B and \textbf{2.8}$\times$ on Qwen3-32B, surpassing the latest methods such as EAGLE-3~\citep{li2025eagle} even with sophisticated training.
\end{itemize}

\section{Background}

\subsection{Speculative Decoding}   \label{sec:prelim1}
Speculative Decoding (SD) operates on a \textit{draft-then-verify} paradigm. A draft model $\mathcal{M}_q$ autoregressively generates $\gamma$ tentative tokens $x_1, \dots, x_\gamma$, which are subsequently evaluated by the target model $\mathcal{M}_p$ in a single parallel forward pass. The acceptance probability for the $i$-th token is defined as:
$
\small
\alpha_i = \min\left(1, \frac{p_{i-1}(x_i)}{q_{i-1}(x_i)}\right).
$
If a token is rejected at $x_i$, a correction is sampled from the distribution $\text{norm}(\max(0, p_{i-1} - q_{i-1}))$; otherwise, if all candidates are accepted, an extra token is appended. 

\noindent\textbf{Theorem 1 (Single-Round)} The expected token generated in a single round is $\mathbb{E}[L_s]$~\citep{leviathan2023fast}:
\begin{equation}
\small
\mathbb{E}[L_s] =  \frac{1 - \alpha^{\gamma+1}}{1 - \alpha}
\end{equation}
\subsection{Retrieval Speculative Decoding} \label{sec:prelim2}

Retrieval-based SD accelerates the target model by substituting autoregressive drafting with retrieved sequences~\citep{he2023rest}. Specifically, given $\bm{x}$, the target model $\mathcal{M}_p$ searches $n$-gram matched tokens and then retrieves $d$ (retrieval depth) tokens from a datastore $\mathcal{D}_R$ as draft tokens. Then $\mathcal{M}_p$ verifies these tokens to yield $s$ matched tokens and one correction, formalized as the \textbf{Retrieval Forward}:
\begin{equation}
\small
\textsc{Retrieval}(\bm{x}, n) = \{x_1, x_2, \dots, x_s, x_{s+1}\}
\end{equation}
Here, $s$ denotes the matched length, ensuring adherence to the target distribution. We define the Average Matched Tokens (AMT) as $\mathbb{E}[s]$, which dictates the effective speedup. Crucially, while existing methods~\citep{token-recycle-luo-2024,pld-saxena-2023} vary in $\mathcal{D}_R$, they share a common oversight: focusing solely on accelerating the target model while neglecting the potential of the draft model.

\section{Theoretical Analysis and Motivation}

\subsection{Speedup Ceiling Theorem}
\label{sec:speed ceiling}
To understand why existing PSD methods plateau, we extend the analysis to a multi-round perspective to unify SD and PSD, formalizing the speedup ceiling that necessitates retrieval-based acceleration.

\paragraph{Definitions} Let $T_p$ and $T_q$ be single-token decoding times for target and draft models, with speed ratio $C = T_p / T_q$. Unlike Theorem 1 that does not consider SD and PSD jointly, we unify them by modeling the process as $k-1$ rounds of continusly accepted tokens followed by a rejection in the $k$-th round, which approximates the bimodal acceptance distribution~\citep{shen2025speculative}.

\paragraph{Theorem 2 (Multi-Round)} The expected number of tokens generated over $k$ rounds is:
\begin{equation}
\small
\mathbb{E}[L_k] = (k-1)(\gamma+1) + \frac{1 - \alpha^{\gamma+1}}{1 - \alpha}
\end{equation}
\noindent\textbf{Theorem 3 (PSD Speedup Ceiling)} The theoretical speedup $S_{\text{PSD}}$ strictly dominates $S_{SD}$ but is upper bounded by $C$:
\begin{equation}
\small
S_{\text{SD}} \le S_{\text{PSD}} = \frac{(\mathbb{E}[L_k]-k +1) \cdot C}{k \cdot \gamma + C} \le \frac{k \cdot\gamma \cdot C}{k \cdot\gamma + C} \le C
\end{equation}
\noindent \textit{Proof.} \quad Within one verification step $T_p$, an autoregressive draft model produces at most $\lfloor T_p / T_q \rfloor \approx C$ tokens. Even with $\alpha=1$, PSD throughput cannot exceed the draft model's generation rate. Formal proofs are available in the Appendix~\ref{sec:appendix_theoretical}.

\begin{figure*}
    \centering
    \begin{subfigure}[b]{0.48\textwidth}  
        \centering
        \includegraphics[width=\textwidth]{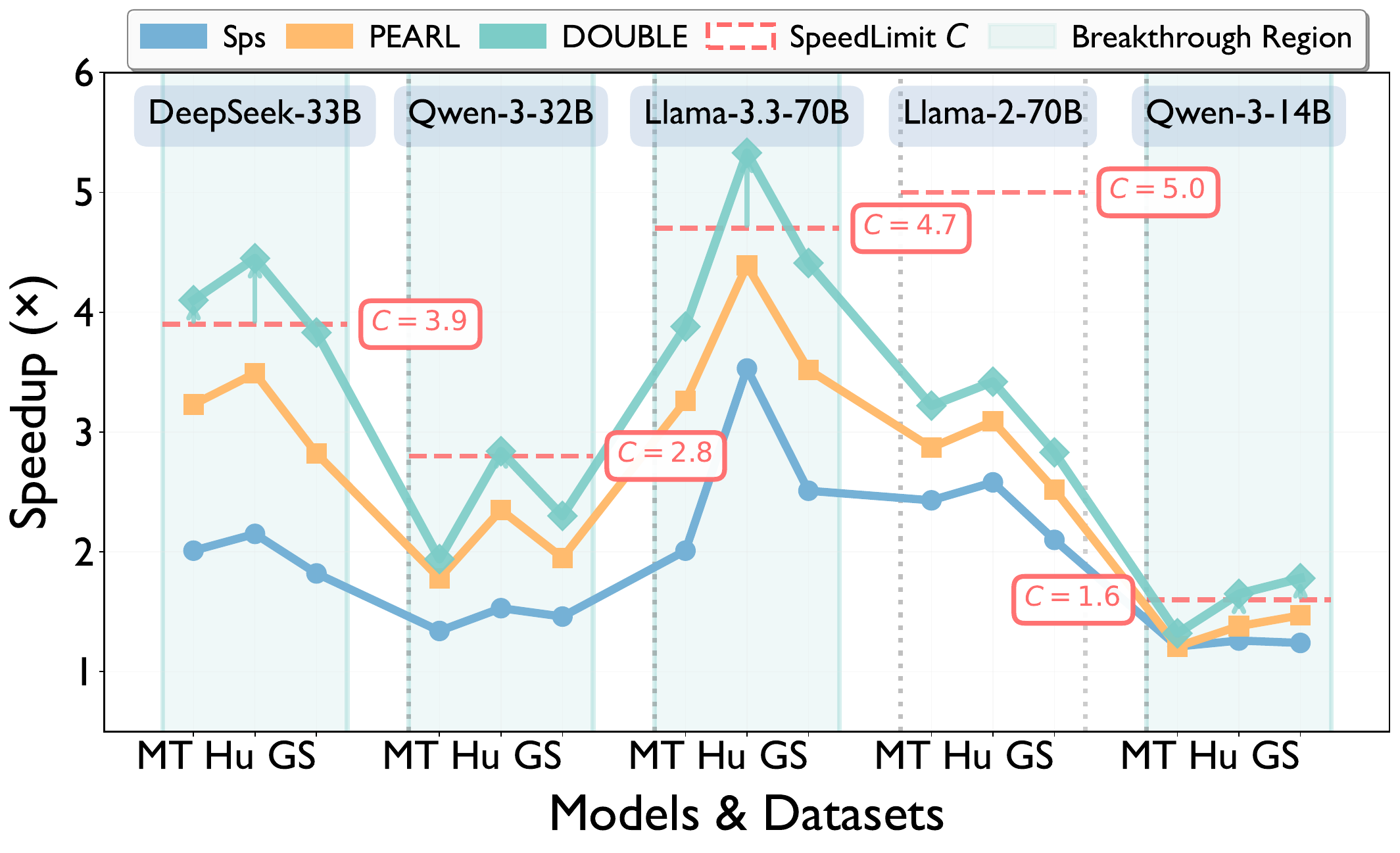}
    \end{subfigure}
    \hspace{0.02\textwidth}  
    \begin{subfigure}[b]{0.48\textwidth}  
        \centering
        \includegraphics[width=\textwidth]{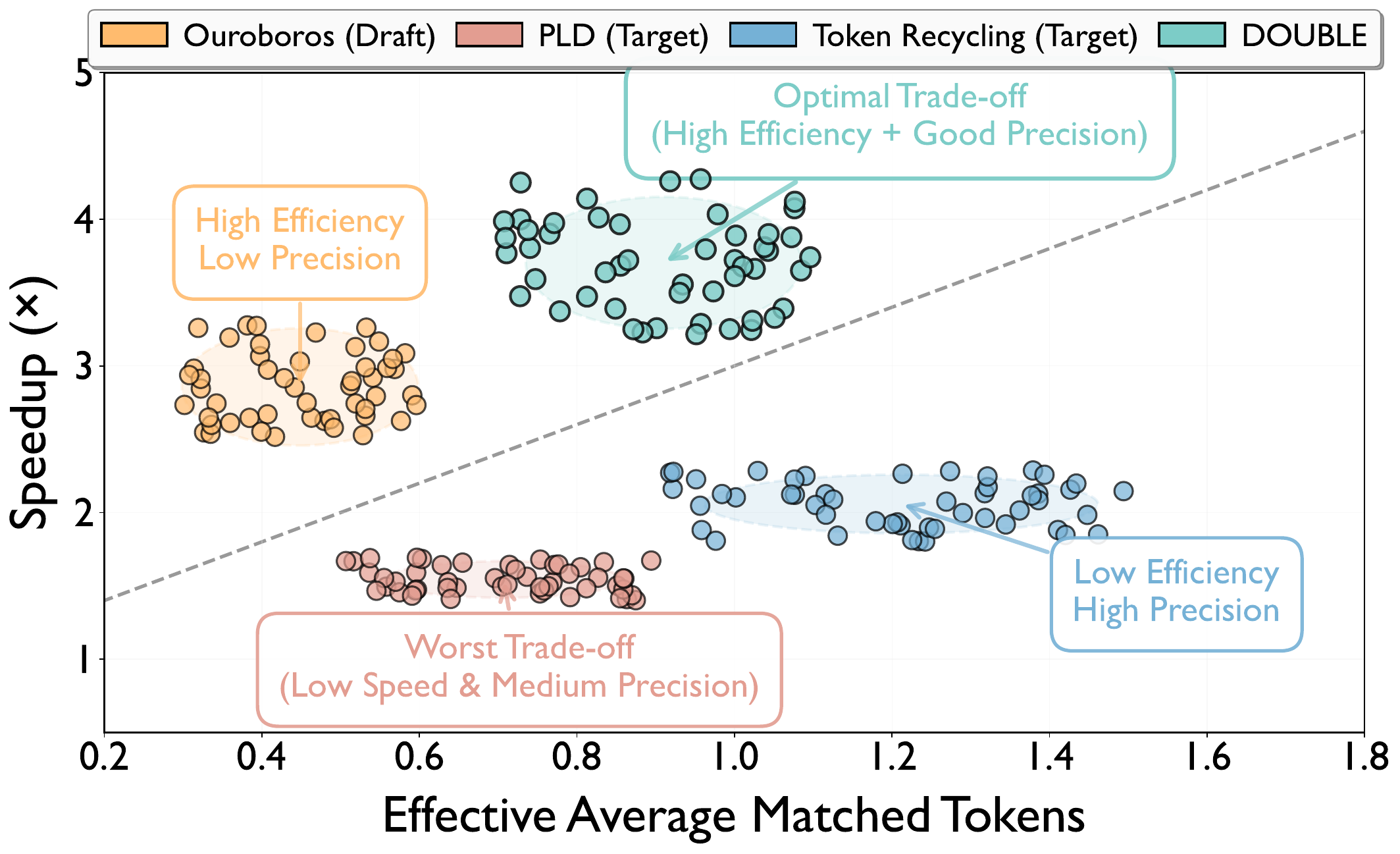}
    \end{subfigure}
    \vspace{-0.1in}
    \caption{Motivation of \textsc{Double}. (a) Breaking the theoretical speedup ceiling $C$ across five model pairs on three benchmarks. Green regions indicate where \textsc{Double} surpasses the speedup limit. (b) Retrieval precision-efficiency trade-off comparison on Deepseek-1.3B\&33B, showing \textsc{Double} achieves optimal balance between effective matched tokens and speedup compared to draft-side (Ouroboros) and target-side (PLD, Token Recycling) methods.}

    \label{fig:motivation}
    \vspace{-0.2in}
\end{figure*}

Theorem 3 establishes a hard ceiling: standard PSD with off-the-shelf models (e.g., $C \approx 2-5$) cannot exceed this factor regardless of acceptance rate. Fig.~\ref{fig:motivation} confirms that both SD and PSD consistently trail this limit. Breaking this barrier requires decoupling draft length from draft latency.


\subsection{Why Double Retrieval?}
\emph{How can we overcome the autoregressive barrier of the draft model to maximize the effective speed ratio $C$?} While Retrieval-based SD proposes reusing tokens to skip autoregression steps~\citep{rest-he-2024}, blindly applying existing methods to PSD exposes a fundamental \textbf{Retrieval Precision-Efficiency Dilemma} as shown in Fig.~\ref{fig:motivation}(b).

\paragraph{\circnumpink{1} Target-Side Retrieval (High Precision, Low Efficiency)} 
Methods like Token Recycling~\citep{token-recycle-luo-2024} only leverage the target model to retrieve. Although accurate (high acceptance), they are bottlenecked by the inherent latency of the target model and verification overhead as shown in the bottom-right of Fig.~\ref{fig:motivation}, thereby trapping these methods in a ``Low Efficiency'' regime.

\paragraph{\circnumpink{2} Draft-Side Retrieval  (High Efficiency, Low Precision)}  Ouroboros~\citep{zhao2024ouroboros} is the only method to shift retrieval to the draft model. Unfortunately, it faces two fundamental limitations: 1) Small draft models lack semantic understanding and capacity for accurate retrieval. The candidates fail verification frequently, resulting in low \textit{effective} AMT after target verification despite long retrieval lengths. This exacerbates computational waste from \textbf{mid-sequence token rejection}~\citep{shen2025speculative}; 2) Ouroboros performs purely serial execution that fails to eliminate the mutual waiting bottleneck between models.
\vspace{-0.05in}

\paragraph{Motivation}

The goal is to achieve speed at draft-side retrieval of generating long chains ($> C$) and the precision of target-side retrieval to prevent those chains from being rejected. We propose \textsc{Double} to fill the precision-efficiency gap that runs retrieval on both draft/target sides simultaneously as detailed in the next section. 

\section{Methodology}
\begin{figure*}[t]
\centering
\includegraphics[width=0.9\linewidth]{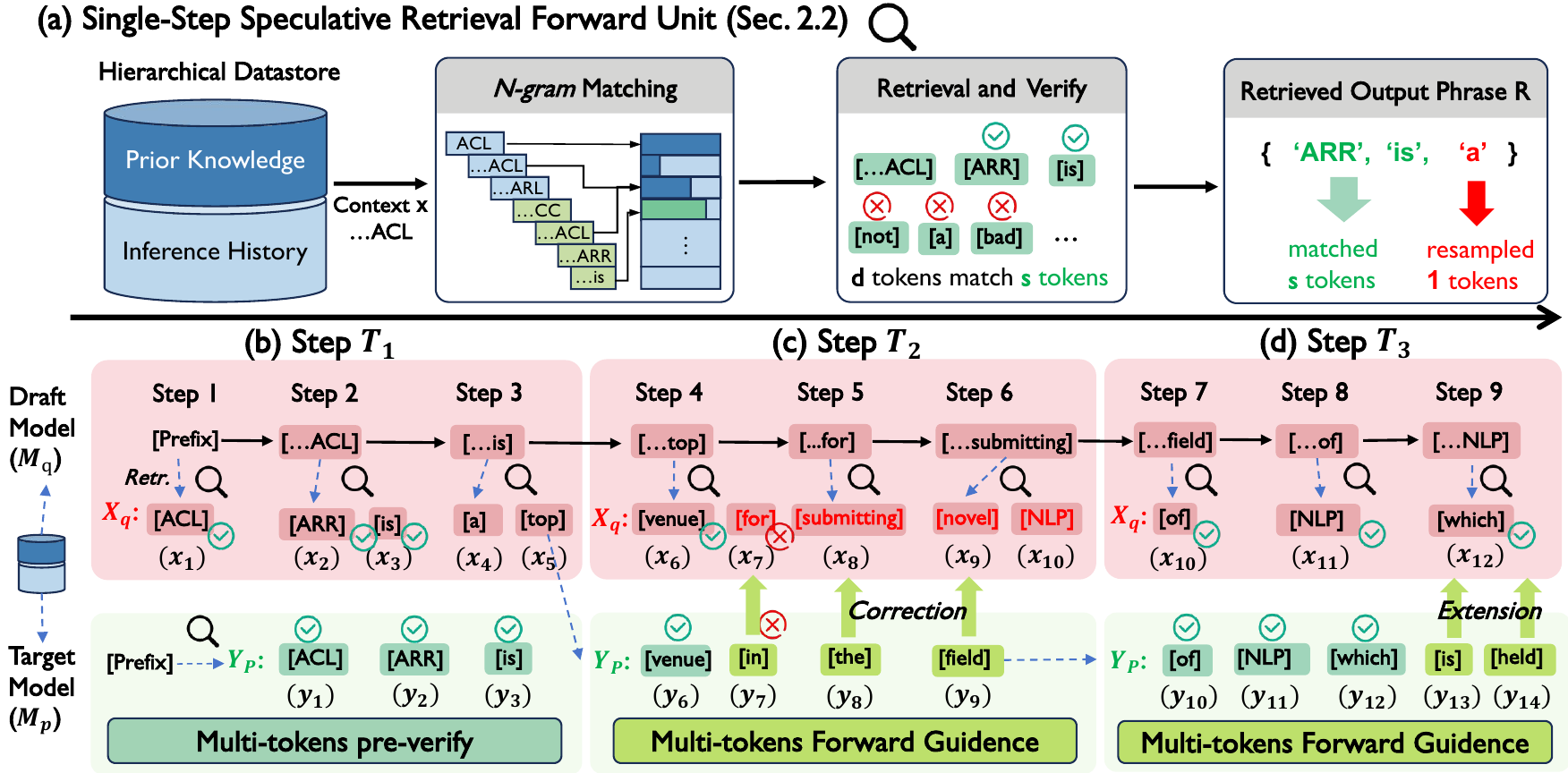}
\vspace*{-0.05in}
\caption{\textbf{Workflow of \textsc{Double}.} 
\textbf{(a) Retrieval Unit:} Utilizes hierarchical datastores to propose $d$ candidates with match length $s$. 
\textbf{(b) Step $T_1$:} At speed ratio $C=3$, $\mathcal{M}_q$ executes iterative retrieval to draft 5 tokens, while $\mathcal{M}_p$ provide the multi-tokens pre-verify. 
\textbf{(c) Step $T_2$:} Target retrieval rectifies $x_{7\text{-}9}$ (``for submitting novel'' $\to$ ``in the field'') as a \textit{Correction}. 
\textbf{(d) Step $T_3$:} Target retrieval directly extends the sequence beyond the draft as an \textit{Extension}.}
\label{fig:framework}
\vspace*{-0.15in}
\end{figure*}

\vspace{-0.05in}
To break the speedup ceiling, we introduce \textsc{Double} (Fig.~\ref{fig:framework}), which consists of: 1) an iterative retrieval drafter that generates the candidate chains; (2) a parallel retrieval architecture that hides latency, and (3) a target-guided verification that turns potential rejections into forward extensions.
\vspace{-0.05in}
\subsection{Parallel Retrieval Decoding}
\label{sec:parallel_retrieval}


\paragraph{Iterative Retrieval Drafter}   
To transcend the theoretical bound $C$, the draft model must generate more than $C$ tokens with a single target forward pass $T_p$.  Instead of standard autoregressive sampling of one token $x \sim q(\cdot|\bm{x})$, the draft model executes $\gamma$ iterations of \emph{Retrieval Forward} defined in Sec.~\ref{sec:prelim2} as shown in (Fig.~\ref{fig:framework}(a)). 
Given context $\bm{x}^{(0)}$, in the $j$-th step iteration ($1 \le j \le C$), the draft model retrieves phrase $R_j$ from the datastore,
\begin{equation}
\small
\vspace{-0.03in}
\mathcal{R}_j = \textsc{Retrieval}(\bm{x}^{(j-1)}, d) = \{x^{(j)}_1, \dots, x^{(j)}_{s_{j+1}}\},
\vspace{-0.03in}
\end{equation}
where $s_j$ represents the number of matched tokens in the $j$-th step, ensuring identical to the original distribution. Then we append $R_j$ to the context and update for the next iteration $\bm{x}^{(j)} = \bm{x}^{(j-1)} \oplus \mathcal{R}_j$. 


\emph{Why retrieval helps break this limit?} In the time it takes an autoregressive draft model to generate $C$ single tokens, the retrieval drafter generates $C$ phrases. For example (Fig.~\ref{fig:framework}(b)), retrieving ``\textit{ARR is}'' ($j=2$) immediately updates the context to retrieve ``\textit{a top}'' ($j=3$).  If the average matched phrase length is \text{AMT}, the total draft length becomes
$L_{\text{draft}} \approx C \times (1 + \text{AMT}) \gg C$,
which breaks the limit and further pushes the effective proposal length beyond the theoretical speed ratio. 

Notably, this sequential update forms a chain structure. Unlike tree-based SD methods that require complex tree attention masks for verification, our linear design remains compatible with standard attention mechanisms. This simplicity eliminates overhead, facilitating flexible deployment and high-concurrency scenarios.

\vspace{-0.1in}
\paragraph{Synchronous Parallel Execution}
We employ a \textit{draft-when-verify} paradigm (Fig.~\ref{fig:framework}(c)) to eliminate pipeline bubbles. While the target model $\mathcal{M}_p$ verifies the candidate sequence from the previous round $k-1$, the draft model $\mathcal{M}_q$ constructs the candidate chain for the next round $k$. This synchronous process fully hides the verification latency, as $\mathcal{M}_q$ prepares candidates of length $L_{\text{draft}} \gg C$ in parallel, effectively resolving the mutual waiting bottleneck.

\vspace{-0.05in}
\subsection{Target-Guided Verification}
\label{sec:double_retrieval}

The previous design only expands the draft length but does not handle the low precision of draft-side retrieval. E.g., in standard PSD, if the draft model makes a mistake at token $x_7$, PSD rejects $x_7,\cdots,x_{N}$ and the computational efforts are wasted. To mitigate such rollbacks, we propose a target-guided verification: the target model $M_p$ performs a single-step retrieval forward to verify the next round unverified tokens and fetch high-confidence $Y_p = \textsc{Retrieval}(\bm{x}, d) = \{y_1, \dots, y_{s+1}\}$ to serve as multi-token pre-verify and forward guidance as detailed below. 
\vspace{-0.05in}
\paragraph{\circnumpink{1} Multi-token Pre-verify}
We employ speculative sampling to validate $X_q$ against the distribution of $Y_p$. Unlike rigid equality checks, this process dynamically identifies a valid length $i$ based on the acceptance criteria and allows divergent draft paths to be filtered out early before stalling the pipeline.
\vspace{-0.05in}
\paragraph{\circnumpink{2} Forward Guidance}
We illustrate forward guidance via an example first. Consider a scenario where the draft model predicts a \emph{low-quality} token at position $7$ ($x_7 = \text{``for''}$). For standard PSD, the target model calculates $p(x_7)$, rejects ``for'', and samples a new token, but this causes a pipeline stall. In contrast, in our framework, the target model retrieves a chain with higher quality, e.g., $Y_p = \{ y_6 = \text{``venue''}, y_7 = \text{``in''}, y_8 = \text{``the''}, y_9 = \text{``field''}\}$. By matching the draft chain $X_q$ against the target guide $Y_p$, we detect a mismatch at $x_7$. Instead of simple rejection, we swap the erroneous $x_7$ with the target's prediction $y_7$, and append the remaining retrieved tokens to the final output, maintaining the original distribution under greedy sampling. 

In short, the target sequence $Y_p$ acts as an active guidance to extend the generation. Formally, we construct the final output sequence by concatenating the verified draft prefix with the remainder of the target sequence. This operation unifies \emph{correction} (Fig.~\ref{fig:framework}(c): $y_i$ mismatch $x_i$) and \emph{extension} (Fig.~\ref{fig:framework}(d): $s+1 > L_{\text{draft}}$) into a single step:
\vspace{-0.05in}
\begin{equation}
\small
x_{\text{next}} = \{x_1, \dots, x_{i-1}\} \oplus \{y_i, \dots, y_{s+1}\}
\vspace{-0.05in}
\end{equation}
Here, the suffix $\{y_{i}, \dots, y_{s+1}\}$ can be considered as a pre-verified \emph{bonus}, as the target model is no longer a passive verifier, but serves as an active guidance for the draft model with fast retrieval. 



\vspace{-0.05in}
\subsection{Hierarchical Datastore Construction}

Existing retrieval methods also face a trade-off in datastore construction challenge. Large-scale external datastores (e.g., REST) incur high memory and training costs~\citep{rest-he-2024}, while on-the-fly datastores (e.g., PLD) suffer from cold-start problems of context sparsity during early inference~\citep{pld-saxena-2023}. Thus, we propose a hierarchical datastore strategy.
\vspace{-0.05in}
\paragraph{Prior Knowledge Initialization}To mitigate the cold-start problem without the latency of external retrieval, we pre-populate the datastore with a compact $n$-gram index derived from limited inference on generic corpora (ShareGPT~\citep{zheng2023judging}). Our analysis (Table~\ref{tab:prior_retrieval}) demonstrates that this lightweight prior suffices to provide robust cross-domain generalization, establishing an immediate retrieval basis with negligible memory cost.
\vspace{-0.25in}
\paragraph{Hybrid Datastore Integration}We construct a hybrid datastore that combines static contexts with dynamic inference history to optimize retrieval coverage. Instead of separate indices, such a datastore is shared by both draft and target models to eliminate redundancy and synchronization overhead. It is initialized with the static prior and user prompt to resolve early-stage sparsity. During decoding, tokens verified by the target model are prioritized, but the rejected tokens are also fused. This ensures retrieval with high quality and diversity.

\vspace{-0.1in}
\subsection{Theoretical Speedup Analysis}
Recall that standard PSD is bounded by $S_{\text{PSD}} \le C$. \textsc{Double} breaks this barrier by decoupling the number of draft tokens from the number of forward passes, which effectively scales up the speed ratio to $C' = C(1 + \text{AMT})$ via retrieval-based generation. Incorporating the draft gain $\mathbb{E}[L_d]$ and target guidance $\mathbb{E}[L_{\text{bonus}}]$, the speedup becomes:
\begin{equation}
\small
    S_{\textsc{Double}} = \frac{(\mathbb{E}[L_d] + \mathbb{E}[L_{\text{bonus}}]) \cdot C(1 + \text{AMT})}{k \cdot\gamma + C(1 + \text{AMT})}
\end{equation}
This demonstrates that \textsc{Double} can break the standard limit and elevate the new theoretical upper bound to a new level of $C(1 + \text{AMT})$.

\vspace{-0.05in}
\section{Experiments}
\label{sec:experiments}
\vspace{-0.05in}
\subsection{Experimental Setting}
\paragraph{Tasks and Datasets}
We evaluate \textsc{Double} with a diverse suite of LLM configurations, including LLaMA-2 (7B/70B)~\citep{grattafiori2024LLaMA3herdmodels}, LLaMA-3 (8B/70B)~\citep{llama3-dubey-2024}, Deepseek-Coder (1.3B/33B)~\citep{deepseek-coder}, and Qwen3 (0.6B/14B/32B)~\citep{yang2025qwen3}. Our evaluation spans five benchmarks covering code, math, summarization, and chat: HumanEval~\citep{chen2021codex}, GSM8K~\citep{Cobbe_Kosaraju_Bavarian_Hilton_Nakano_Hesse_Schulman_2021}, CNN/DM~\citep{Nallapati_Zhou_dos}, Alpaca~\citep{taori2023alpaca}, and MT-Bench~\citep{mt-bench-zheng-2023}. 

\vspace{-0.05in}
\paragraph{Baselines and Implementation}
We benchmark against representative methods across different SD categories: 
1) \textbf{Standard SD}~\citep{chen2023accelerating}; 
2) \textbf{Target-side Retrieval} (Lookahead~\citep{fu2024break}, PLD~\citep{pld-saxena-2023}, Token Recycling~\citep{token-recycle-luo-2024}); 
3) \textbf{Draft-side Retrieval} (Ouroboros~\citep{zhao2024ouroboros}); 
4) \textbf{Parallel SD} (PEARL~\citep{liu2024parallel}); and 
5) the SOTA \textbf{Training-based} method, EAGLE-3~\citep{li2025eagle}. 
Experiments are conducted on 8 NVIDIA A100 (80GB) GPUs under greedy sampling. We report Wall-time Speedup and Mean Accepted Tokens ($M$), where $M$ denotes the continuously accepted length for parallel frameworks~\citep{liu2024parallel}. We set the draft length $\gamma = \lceil C \rceil$, with $C$ representing the average speed ratio across datasets. Details are in Appendix D.1.

\begin{figure}[t]
\includegraphics[width=\linewidth]{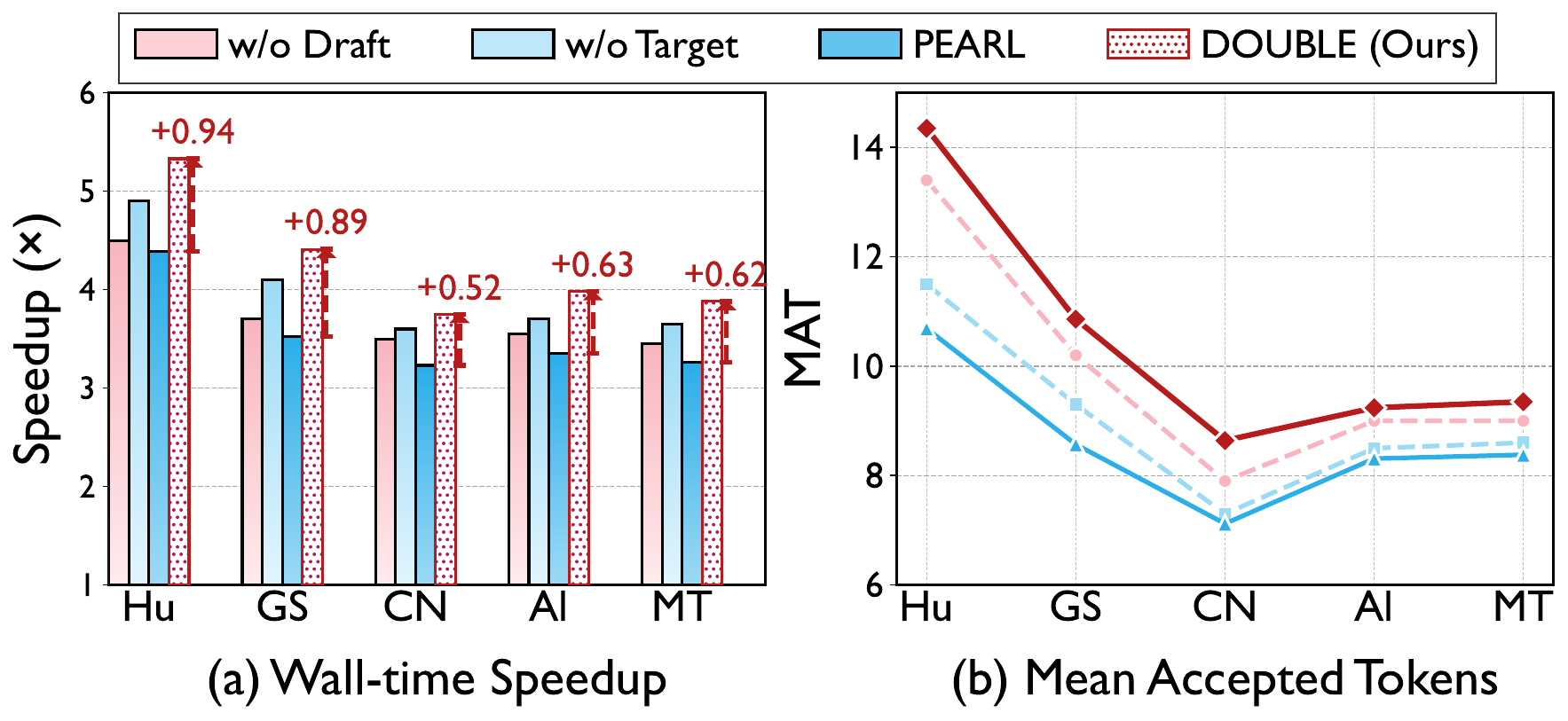}
\vspace{-0.25in}
\caption{Ablation study on LLaMA-3.3-70B: both retrieval components are indispensable for achieving optimal speedup and accepted length.}
\label{ablation_component}
\vspace{-0.2in}
\end{figure}

\subsection{Main Results}
\label{sec:main_results}
Table~\ref{tab:main_resultes} demonstrates that \textsc{Double} achieves consistent speedups ranging from $1.6\times$ to $5.3\times$ across five benchmarks, highlighting three key advantages:
\noindent\textbf{(\uppercase\expandafter{\romannumeral1}) Breaking the PSD Speed Ceiling.} 
Existing parallel frameworks (e.g., PEARL) are strictly bounded by the draft-to-target speed ratio $C$. By decoupling draft length from latency, \textsc{Double} breaks this theoretical limit. This is evident on the constrained Qwen3-14B pair ($C \approx 1.6$), where \textsc{Double} achieves a $1.89\times$ speedup on CNN/DM, surpassing the inherent ceiling and outperforming PEARL ($1.28\times$) by over $47\%$.
\noindent\textbf{(\uppercase\expandafter{\romannumeral2}) Tackling the Precision-Efficiency Dilemma.} 
\textsc{Double} effectively overcomes the limitations of single-sided retrieval. Compared to Target-side methods (e.g., TR) limited by verification overhead, \textsc{Double} drastically boosts throughput (e.g., improving Deepseek-33B speedup from $1.96\times$ to $4.05\times$). Conversely, different from Draft-side only methods (e.g., Ouroboros) that are plagued by low precision, our target-guided mechanism ensures high acceptance. This leads to $3.42\times$ on LLaMA2-70B compared to Ouroboros's $2.08\times$.
\noindent\textbf{(\uppercase\expandafter{\romannumeral3}) Advantages over Training-Based Methods.} 
Despite \textit{zero} training, \textsc{Double} matches or exceeds the SOTA training-based method, EAGLE-3. On models (LLaMA-3.3, Qwen3-32B), it attains speedups of $5.33\times$ and $2.84\times$ on HumanEval, respectively. This validates our strategy yields acceleration that can often exceed expensive architectural modifications, offering a promising alternative to EAGLE-3.

\begin{table*}[t]
\caption{\textbf{Main results on five benchmarks.} Performance comparison between \textsc{Double} and existing baselines across diverse model configurations. Note that LLaMA-3.1-8B serves as the draft model for LLaMA-3.3-70B. \textbf{Bold} numbers denote the best performance and \underline{Underline} denotes when \textsc{Double} breaks the speed limit.}
\label{tab:main_resultes}
\centering
\resizebox{\linewidth}{!}{%
\begin{tabular}{c l cc cc cc cc cc c}
\toprule
\multirow{2}{*}{Models} & \multirow{2}{*}{Methods} & \multicolumn{2}{c}{HumanEval} & \multicolumn{2}{c}{GSM8K} & \multicolumn{2}{c}{CNN/DM} & \multicolumn{2}{c}{Alpaca} & \multicolumn{2}{c}{MT-Bench}&\multirow{2}{*}{Avg.}\\
\cmidrule(lr){3-4} \cmidrule(lr){5-6} \cmidrule(lr){7-8} \cmidrule(lr){9-10} \cmidrule(lr){11-12} & &M & Speedup & M & Speedup & M & Speedup & M & Speedup & M & Speedup\\
\midrule
\multirow{6}{*}{\makecell[c]{Deepseek-33B\\(C $\approx$ 3.9)} }
& Lookahead & 2.34 & 1.75$\times$ &  1.83&  1.45$\times$ &  1.87&  1.66$\times$ &  1.69&  1.34$\times$ &  1.65 & 1.36$\times$ &  1.51$\times$\\
& PLD   & 1.86 & 1.68$\times$ &  1.84&  1.62$\times$ &  2.07&  1.87$\times$ &  1.89&  1.74$\times$ &  1.69 & 1.47$\times$ &  1.68$\times$\\
& TR & 2.83&  2.14$\times$&  2.67&  1.98$\times$&  2.88& 2.17$\times$ &  2.45&  1.83$\times$&  2.36& 1.72$\times$&  1.96$\times$\\
& Ouroboros & 6.38& 3.15$\times$  &  5.95&  2.88$\times$ & 5.97 & 2.92$\times$  & 5.30 & 2.64$\times$  & 5.12 & 2.36$\times$& 2.79$\times$\\
& Sps &  5.09& 2.15$\times$&  3.94& 1.82$\times$ &  4.18 & 2.05$\times$ & 4.22 & 2.16$\times$ & 3.92& 2.01$\times$& 2.04$\times$  \\
& PEARL & 14.51&  3.49$\times$&  7.83&  2.82$\times$& 7.76 & 2.77$\times$ &  8.71&  2.86$\times$& 7.92 & 3.23$\times$& 3.03$\times$\\
&\cellcolor{pink!30}\textbf{DOUBLE} & \cellcolor{pink!30}\textbf{16.47} & \cellcolor{pink!30} \underline{\textbf{4.45$\times$}}& \cellcolor{pink!30}\textbf{10.86} & \cellcolor{pink!30}\textbf{3.83$\times$} & \cellcolor{pink!30}\textbf{10.64}  & \cellcolor{pink!30}\underline{\textbf{4.05$\times$}}&
\cellcolor{pink!30}\textbf{10.24} & \cellcolor{pink!30}\textbf{3.84$\times$}&
\cellcolor{pink!30}\textbf{9.85} & \cellcolor{pink!30}\underline{\textbf{4.10$\times$}}&
\cellcolor{pink!30}\underline{\textbf{4.05$\times$}}   \\

&\cellcolor{blue1}{$\bigtriangleup$ $(\uparrow, \%)$} & \cellcolor{blue1}{$\uparrow$ 13.5\%} & \cellcolor{blue1} {\textcolor{red}{$\uparrow$ 27.5\%}}& \cellcolor{blue1}{$\uparrow$ 38.7\%} & \cellcolor{blue1}{\textcolor{red}{$\uparrow$ 35.8\%}} & \cellcolor{blue1}{$\uparrow$ 37.1\%}  & \cellcolor{blue1}{\textcolor{red}{$\uparrow$ 46.2\%}}&
\cellcolor{blue1}{$\uparrow$ 17.6\%} & \cellcolor{blue1}{\textcolor{red}{$\uparrow$ 34.3\%}}&
\cellcolor{blue1}{$\uparrow$ 24.4\%} & \cellcolor{blue1}{\textcolor{red}{$\uparrow$ 26.9\%}}&
\cellcolor{blue1}{\textcolor{red}{$\uparrow$ 33.7\%}}  \\

\midrule
\multirow{6}{*}{\makecell[c]{LLaMA2-70B\\(C $\approx$ 5.0)} }
& Lookahead & 1.86 & 1.58$\times$ &  2.04&  1.62$\times$ &  1.57&  1.27$\times$ &  1.69&  1.31$\times$ &  1.72 & 1.45$\times$ &  1.45$\times$\\
& PLD   & 1.66 & 1.53$\times$ &  1.64&  1.52$\times$ &  1.97&  1.77$\times$ &  1.67&  1.54$\times$ &  1.49 & 1.45$\times$ &  1.57$\times$\\
& TR & 2.43&  2.02$\times$&  2.27&  1.98$\times$&  2.58& 2.21$\times$ &  1.86&  1.73$\times$&  1.75& 1.67$\times$&  1.92$\times$\\
& Ouroboros & 5.08& 2.08$\times$  &  4.95&  1.97$\times$ & 5.21 & 2.25$\times$  & 4.90 & 1.84$\times$  & 4.82 & 1.76$\times$& 1.98$\times$\\
& Sps &  5.42& 2.58$\times$&  4.34& 2.10$\times$ &  4.18 & 2.03$\times$ & 4.22 & 2.06$\times$ & 4.32& 2.43$\times$& 2.24$\times$  \\
& PEARL &  7.24&  3.09$\times$&  6.83&  2.52$\times$& 6.76 & 2.37$\times$ &  7.71&  2.66$\times$& 6.98 & 2.87$\times$& 2.70$\times$\\
&\cellcolor{pink!30}\textbf{DOUBLE} & \cellcolor{pink!30}\textbf{8.65} & \cellcolor{pink!30} \textbf{3.42$\times$}& \cellcolor{pink!30}\textbf{7.86} & \cellcolor{pink!30}\textbf{2.83$\times$} & \cellcolor{pink!30}\textbf{7.84}  & \cellcolor{pink!30}\textbf{3.05$\times$}&
\cellcolor{pink!30}\textbf{8.24} & \cellcolor{pink!30}\textbf{2.94$\times$}&
\cellcolor{pink!30}\textbf{8.35} & \cellcolor{pink!30}\textbf{3.22$\times$}&
\cellcolor{pink!30}\textbf{3.09$\times$}   \\

&\cellcolor{blue1}{$\bigtriangleup$ $(\uparrow, \%)$} & \cellcolor{blue1}{$\uparrow$ 19.5\%} & \cellcolor{blue1} {\textcolor{red}{$\uparrow$ 10.7\%}}& \cellcolor{blue1}{$\uparrow$ 15.1\%} & \cellcolor{blue1}{\textcolor{red}{$\uparrow$ 12.3\%}} & \cellcolor{blue1}{$\uparrow$ 16.0\%}  & \cellcolor{blue1}{\textcolor{red}{$\uparrow$ 28.7\%}}&
\cellcolor{blue1}{$\uparrow$ 6.9\%} & \cellcolor{blue1}{\textcolor{red}{$\uparrow$ 10.5\%}}&
\cellcolor{blue1}{$\uparrow$ 19.6\%} & \cellcolor{blue1}{\textcolor{red}{$\uparrow$ 12.2\%}}&
\cellcolor{blue1}{\textcolor{red}{$\uparrow$ 14.4\%}}  \\

\midrule
\multirow{5}{*}{\makecell[c]{LLaMA3.3-70B\\(C $\approx$ 4.7)} }
& PLD   & 1.76 & 1.58$\times$ &  1.54&  1.42$\times$ &  1.87&  1.67$\times$ &  1.69&  1.54$\times$ &  1.59 & 1.46$\times$ &  1.53$\times$\\
& Sps  & 5.40 & 3.53$\times$ &  4.97&  2.51$\times$&  4.83&  2.46$\times$&  4.61&  2.23$\times$&  4.30& 2.01$\times$& 2.55$\times$\\
& EAGLE3 & 6.51 & 4.56$\times$ & 6.02 & 4.21$\times$ & 4.82 & 3.03$\times$ & 5.89 &  \textbf{4.13}$\times$&  5.23& \textbf{4.02}$\times$&  3.99$\times$\\
& PEARL & 10.69 & 4.39$\times$ &  8.57&  3.52$\times$&  7.12& 3.23$\times$& 8.31 & 3.35$\times$ & 8.38& 3.26$\times$& 3.55$\times$ \\
&\cellcolor{pink!30}\textbf{DOUBLE} & \cellcolor{pink!30}\textbf{14.35} & \cellcolor{pink!30} \underline{\textbf{5.33$\times$}}& \cellcolor{pink!30}\textbf{10.86} & \cellcolor{pink!30}\textbf{4.41$\times$} & \cellcolor{pink!30}\textbf{8.64}  & \cellcolor{pink!30}\textbf{3.75$\times$}&
\cellcolor{pink!30}\textbf{9.24} & \cellcolor{pink!30}3.98$\times$&
\cellcolor{pink!30}\textbf{9.35} & \cellcolor{pink!30}3.88$\times$&
\cellcolor{pink!30}\textbf{4.27$\times$}   \\

&\cellcolor{blue1}{$\bigtriangleup$ $(\uparrow, \%)$} & \cellcolor{blue1}{$\uparrow$ 34.2\%} & \cellcolor{blue1} {\textcolor{red}{$\uparrow$ 21.4\%}}& \cellcolor{blue1}{$\uparrow$ 26.7\%} & \cellcolor{blue1}{\textcolor{red}{$\uparrow$ 25.3\%}} & \cellcolor{blue1}{$\uparrow$ 21.3\%}  & \cellcolor{blue1}{\textcolor{red}{$\uparrow$ 16.1\%}}&
\cellcolor{blue1}{$\uparrow$ 11.2\%} & \cellcolor{blue1}{\textcolor{red}{$\uparrow$ 18.8\%}}&
\cellcolor{blue1}{$\uparrow$ 11.6\%} & \cellcolor{blue1}{\textcolor{red}{$\uparrow$ 19.0\%}}&
\cellcolor{blue1}{\textcolor{red}{$\uparrow$ 20.3\%}}  \\

\midrule
\multirow{5}{*}{\makecell[c]{Qwen3-14B\\(C $\approx$ 1.6)} }
& PLD   & 1.52 & 1.38$\times$ &  1.44&  1.31$\times$ &  1.67&  1.53$\times$ &  1.53&  1.35$\times$ &  1.47 & 1.35$\times$ &  1.38$\times$\\
& Sps  & 3.37 & 1.26$\times$ &  3.44&  1.24$\times$&  3.38&  1.13$\times$&  3.21&  1.23$\times$&  3.17& 1.21$\times$& 1.21$\times$\\
& EAGLE3 & 2.99 & \textbf{1.91}$\times$ & 3.02 & \textbf{1.95}$\times$ & 2.41 & 1.58$\times$ & 2.49 &  \textbf{1.65}$\times$&  2.83& \textbf{1.89}$\times$&  \textbf{1.80}$\times$\\
& PEARL & 7.24 & 1.38$\times$ &  7.36&  1.47$\times$&  5.62& 1.28$\times$& 5.76 & 1.31$\times$ & 3.77& 1.21$\times$& 1.33$\times$ \\
&\cellcolor{pink!30}\textbf{DOUBLE} & \cellcolor{pink!30}\textbf{12.65} & \cellcolor{pink!30} \underline{1.65$\times$}& \cellcolor{pink!30}\textbf{10.53} & \cellcolor{pink!30}\underline{1.78$\times$} & \cellcolor{pink!30}\textbf{9.64}  & \cellcolor{pink!30}\underline{\textbf{1.89$\times$}}&
\cellcolor{pink!30}\textbf{8.24} & \cellcolor{pink!30}1.54$\times$&
\cellcolor{pink!30}\textbf{5.13} & \cellcolor{pink!30}1.32$\times$&
\cellcolor{pink!30}\underline{1.63}$\times$   \\

&\cellcolor{blue1}{$\bigtriangleup$ $(\uparrow, \%)$} & \cellcolor{blue1}{$\uparrow$ 74.7\%} & \cellcolor{blue1} {\textcolor{red}{$\uparrow$ 19.6\%}}& \cellcolor{blue1}{$\uparrow$ 43.1\%} & \cellcolor{blue1}{\textcolor{red}{$\uparrow$ 21.1\%}} & \cellcolor{blue1}{$\uparrow$ 71.5\%}  & \cellcolor{blue1}{\textcolor{red}{$\uparrow$ 47.7\%}}&
\cellcolor{blue1}{$\uparrow$ 43.1\%} & \cellcolor{blue1}{\textcolor{red}{$\uparrow$ 17.6\%}}&
\cellcolor{blue1}{$\uparrow$ 36.1\%} & \cellcolor{blue1}{\textcolor{red}{$\uparrow$ 9.1\%}}&
\cellcolor{blue1}{\textcolor{red}{$\uparrow$ 22.6\%}}  \\

\midrule
\multirow{5}{*}{\makecell[c]{Qwen3-32B\\(C $\approx$ 2.8)} }
& PLD   & 1.46 & 1.16$\times$ &  1.41&  1.12$\times$ &  1.56&  1.47$\times$ &  1.49&  1.31$\times$ &  1.43 & 1.29$\times$ &  1.27$\times$\\
& Sps  & 4.27 & 1.53$\times$ &  4.04&  1.46$\times$&  3.38&  1.33$\times$&  3.71&  1.43$\times$&  3.47& 1.34$\times$& 1.42$\times$\\
& EAGLE3 & 2.83 & 2.27$\times$ & 2.99 & \textbf{2.53}$\times$ & 2.57 & 1.73$\times$ & 2.78 & 1.80$\times$ & 3.03 & \textbf{2.23}$\times$ & 2.11$\times$\\
& PEARL & 7.78 & 2.35$\times$ &  7.53&  1.95$\times$&  4.12& 1.72$\times$& 3.86 & 1.83$\times$ & 3.68& 1.78$\times$& 1.92$\times$ \\
&\cellcolor{pink!30}\textbf{DOUBLE} & \cellcolor{pink!30}\textbf{10.09} & \cellcolor{pink!30} \underline{\textbf{2.84$\times$}}& \cellcolor{pink!30}\textbf{8.83} & \cellcolor{pink!30}2.30$\times$ & \cellcolor{pink!30}\textbf{6.64}  & \cellcolor{pink!30}\textbf{2.56$\times$}&
\cellcolor{pink!30}\textbf{6.24} & \cellcolor{pink!30}\textbf{2.04$\times$}&
\cellcolor{pink!30}\textbf{5.83} & \cellcolor{pink!30}1.94$\times$&
\cellcolor{pink!30}\textbf{2.33$\times$}  \\

&\cellcolor{blue1}{$\bigtriangleup$ $(\uparrow, \%)$} & \cellcolor{blue1}{$\uparrow$ 29.7\%} & \cellcolor{blue1} {\textcolor{red}{$\uparrow$ 20.9\%}}& \cellcolor{blue1}{$\uparrow$ 17.3\%} & \cellcolor{blue1}{\textcolor{red}{$\uparrow$ 17.9\%}} & \cellcolor{blue1}{$\uparrow$ 61.2\%}  & \cellcolor{blue1}{\textcolor{red}{$\uparrow$ 48.8\%}}&
\cellcolor{blue1}{$\uparrow$ 61.7\%} & \cellcolor{blue1}{\textcolor{red}{$\uparrow$ 11.5\%}}&
\cellcolor{blue1}{$\uparrow$ 58.4\%} & \cellcolor{blue1}{\textcolor{red}{$\uparrow$ 9.0\%}}&
\cellcolor{blue1}{\textcolor{red}{$\uparrow$ 21.4\%}}  \\

\bottomrule
\end{tabular}
}
\vspace{-0.15in}
\end{table*}

\begin{table*}[t]

\caption{Performance comparison between \textsc{DOUBLE} and existing baseline under non-greedy settings.}
\label{tab:nongreedy}
\centering
\resizebox{\linewidth}{!}{
\begin{tabular}{lcccccccccccc}
\toprule
\multirow{2}{*}{Methods} & \multicolumn{2}{c}{HumanEval} & \multicolumn{2}{c}{GSM8K} & \multicolumn{2}{c}{CNN/DM} & \multicolumn{2}{c}{Alpaca} & \multicolumn{2}{c}{MT-Bench} & \multicolumn{2}{c}{Avg.}\\
\cmidrule(lr){2-3} \cmidrule(lr){4-5} \cmidrule(lr){6-7} \cmidrule(lr){8-9} \cmidrule(lr){10-11} \cmidrule(lr){12-13}
 & $M$ & Speedup & $M$ & Speedup & $M$ & Speedup & $M$ & Speedup & $M$ & Speedup & $M$ & Speedup\\
\midrule
\multicolumn{13}{c}{\cellcolor{cyan!10}LLaMA3.3-70B, \emph{Temperature = 1.0}} \\
\midrule
EAGLE3 & 6.05 & 4.24$\times$ & 5.60 & \textbf{3.91}$\times$ & 4.49 & 2.82$\times$ & 5.48 & \textbf{3.84$\times$} & 4.86 & \textbf{3.74$\times$} & 5.30 & 3.71$\times$\\
PEARL & 10.06 & 4.13$\times$ & 8.06 & 3.31$\times$ & 6.69 & 3.04$\times$ & 7.82 & 3.15$\times$ & 7.89 & 3.07$\times$ & 8.10 & 3.34$\times$ \\
\cellcolor{pink!30}\textbf{DOUBLE} & \cellcolor{pink!30}\textbf{12.66} & \cellcolor{pink!30}\textbf{4.70$\times$} & \cellcolor{pink!30}\textbf{9.58} & \cellcolor{pink!30}3.89$\times$ & \cellcolor{pink!30}\textbf{7.62} & \cellcolor{pink!30}\textbf{3.31$\times$} & \cellcolor{pink!30}\textbf{8.15} & \cellcolor{pink!30}3.51$\times$ & \cellcolor{pink!30}\textbf{8.25} & \cellcolor{pink!30}3.42$\times$ & \cellcolor{pink!30}\textbf{9.25} & \cellcolor{pink!30}\textbf{3.77$\times$} \\
\midrule
\multicolumn{13}{c}{\cellcolor{gray!20}Qwen3-32B, \emph{Temperature = 1.0}} \\
\midrule
EAGLE3 & 2.63 & 2.11$\times$ & 2.78 & \textbf{2.35$\times$} & 2.39 & 1.61$\times$ & 2.58 & 1.67$\times$ & 2.81 & \textbf{2.07$\times$} & 2.64 & 1.96$\times$\\
PEARL & 7.31 & 2.21$\times$ & 7.07 & 1.83$\times$ & 3.86 & 1.62$\times$ & 3.63 & 1.72$\times$ & 3.45 & 1.67$\times$ & 5.06 & 1.81$\times$ \\
\cellcolor{pink!30}\textbf{DOUBLE} & \cellcolor{pink!30}\textbf{8.90} & \cellcolor{pink!30}\textbf{2.50$\times$} & \cellcolor{pink!30}\textbf{7.80} & \cellcolor{pink!30}2.03$\times$ & \cellcolor{pink!30}\textbf{5.86} & \cellcolor{pink!30}\textbf{2.26$\times$} & \cellcolor{pink!30}\textbf{5.51} & \cellcolor{pink!30}\textbf{1.80$\times$} & \cellcolor{pink!30}\textbf{5.15} & \cellcolor{pink!30}1.72$\times$ & \cellcolor{pink!30}\textbf{6.64} & \cellcolor{pink!30}\textbf{2.06$\times$} \\
\bottomrule
\end{tabular}
}
\vspace{-0.15in}
\end{table*}

\subsection{Ablation Study}
\label{sec:ablation}
\paragraph{Component Analysis}
To evaluate the component-wise contribution, we analyze the impact of Draft-side and Target-side retrieval on LLaMA-3.3. As shown in Fig.~\ref{ablation_component}, the results demonstrate that both components are critical:
\begin{itemize}[leftmargin=*, itemsep=0pt, topsep=2pt]
    \item \textbf{Impact of Draft Retrieval:} Removing draft retrieval (\textit{w/o Draft}) reverts the system to autoregressive drafting, effectively constraining speedup to the theoretical ceiling $C$. By enabling iterative retrieval, \textsc{Double} decouples proposal length from latency, elevating the speedup on HumanEval from $4.50\times$ to \textbf{5.33$\times$}.
    \vspace{-0.05in}
    \item \textbf{Impact of Target Retrieval:} Excluding target retrieval (\textit{w/o Target}) degrades precision due to the draft model's limitations, reducing Mean Accepted Tokens (MAT) from $14.35$ to $11.5$. Target guidance functions as a correction filter, saving tokens that are meant to be rejected in order to maintain high acceptance rates.
\end{itemize}
The synergy between them allows us to surpass PEARL ($4.39\times$, $10.69$ MAT), effectively resolving the Precision-Efficiency Dilemma by balancing high speed with robust verification precision.

\begin{figure}[t]
    \centering
    \includegraphics[width=0.48\textwidth]{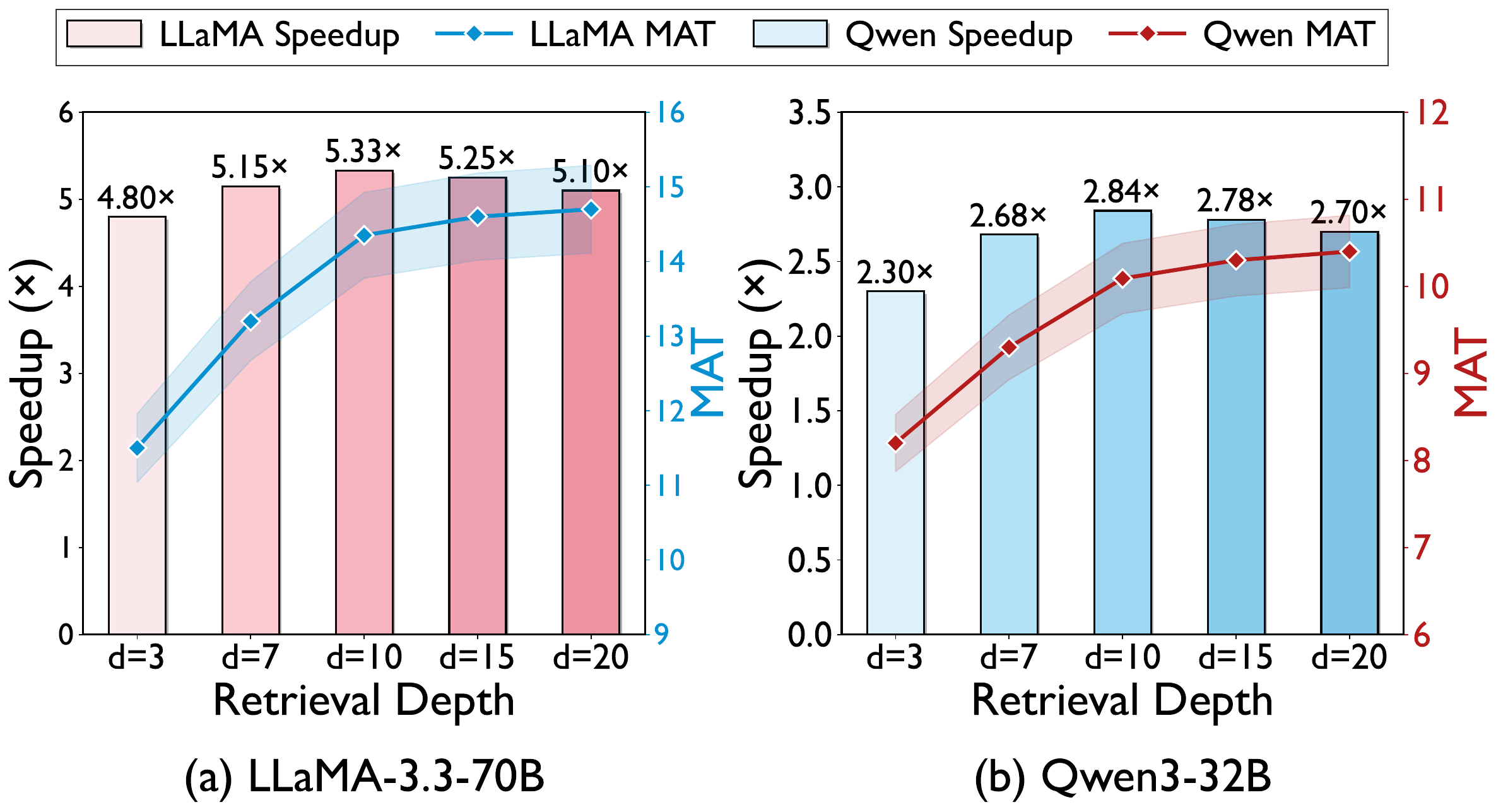}
\vspace{-0.25in}
\caption{Impact of retrieval depth $d$. MAT saturates and speedup fluctuates beyond $d=10$.}
    \label{fig:ablation_depth}
    \vspace{-0.2in}
\end{figure}

\vspace{-0.05in}
\paragraph{Temperature Sampling}
Under non-greedy settings, strictly preserving the target distribution necessitates a modified verification protocol. Upon rejecting a token $x_i$, we sample a correction from the residual distribution $\text{norm}(\max(0, p_{i-1} - q_{i-1}))$. Although distributional consistency mandates discarding subsequent guidance $y_{>i}$, the \textit{Multi-token Pre-verify} mechanism remains active for the prefix, effectively pruning divergent tokens to minimize overhead. Table~\ref{tab:nongreedy} validates performance under stochastic sampling ($T=1.0$): on LLaMA3.3-70B, \textsc{Double} achieves \textbf{4.70$\times$} speedup on HumanEval and averages \textbf{3.77$\times$}, consistently outperforming EAGLE3 ($3.71\times$) and PEARL ($3.34\times$). These results confirm that \textsc{Double} maintains high acceleration efficiency even in stochastic regimes.
\vspace{-0.05in}
\paragraph{Effect of Retrieval Depth}
We analyze the impact of retrieval depth $d$ on efficiency. As shown in Fig.~\ref{fig:ablation_depth}, increasing $d$ initially boosts the Mean Accepted Tokens ($M$), but the performance saturates beyond $d=10$. This is intuitive because multi-step prediction reduces accuracy. As the retrieval chain extends, distant tokens become less relevant to the current context, leading to a lower acceptance rate. Thus, we select $d=10$ as the optimal balance.

\vspace{-0.05in}
\paragraph{Prior Retrieval Knowledge}
Table~\ref{tab:prior_retrieval} analyzes the efficacy of our prior knowledge datastore, serialized as a lightweight local \texttt{.pkl} file. We demonstrate that a single datastore derived from generic ShareGPT data achieves robust generalization across diverse benchmarks (HumanEval, GSM8K, MT-Bench), obviating the need for task-specific adaptation. Performance gains saturate at 10 rounds (9.5 MB); extending initialization to 20 rounds yields diminishing returns while linearly increasing storage overhead. Consequently, we establish $K=10$ as the optimal configuration to balance the overhead.

\vspace{-0.05in}
\paragraph{More Discussion}
Due to space, we include more results and discussions in the Appendix, including
high batch sizes analysis (\ref{sec:appendix_batch}), lossless analysis (\ref{sec:appendix_lossless}), time consumption(\ref{sec:appendix_time}), hierarchical datastore(\ref{sec:appendix_datastore}) and more ablation results (\ref{sec:appendix_dilemma}, \ref{sec:appendix_memory}).

\section{Related Work}
\label{sec:related_work}
\vspace{-0.05in}
\paragraph{Speculative Decoding}
Standard SD accelerates inference by drafting tokens in a lossless manner, yet maximizing the acceptance rate remains a key challenge. Strategies range from training-based auxiliary modules (Medusa~\citep{cai2024medusa}, EAGLE~\citep{li2024eagle}) to training-free verification trees (SpecInfer~\citep{chen2023accelerating}). Although recent advancements like EAGLE-3~\citep{li2025eagle} have optimized small model scaling, these methods adhere to a sequential \emph{draft-then-verify} paradigm, which imposes an unavoidable mutual waiting bottleneck between models.
\vspace{-0.05in}
\paragraph{Parallel Speculative Decoding}
To address serial inefficiencies, PSD frameworks such as PEARL~\citep{liu2024parallel} and SpecBranch~\citep{shen2025speculative} introduce pipelined execution, enabling concurrent drafting and verification. While this effectively utilizes idle compute, the overall speedup remains strictly upper-bounded by the inherent speed ratio between draft and target models. Furthermore, these pipelines are highly sensitive to latency penalties once mid-token rejections occur.

\vspace{-0.05in}
\paragraph{Retrieval Speculative Decoding}
Retrieval-based methods utilize $n$-gram matching but typically isolate acceleration to either the \textbf{Target-Side} (Lookahead~\citep{fu2024break}, PLD~\citep{pld-saxena-2023}) or \textbf{Draft-Side} (Ouroboros~\citep{zhao2024ouroboros}). This isolation creates a fundamental \emph{Precision-Efficiency Dilemma}. In contrast, we propose \textbf{\textsc{Double}}, which orchestrates synergistic retrieval across both models to resolve this trade-off, effectively breaking the theoretical speedup ceiling of existing parallel frameworks.

\begin{table}[t]
    \centering
    \caption{Impact of initialization rounds $K$ on storage and speedup (Qwen3-32B).}
    \label{tab:prior_retrieval}
    \resizebox{\linewidth}{!}{%
    \begin{tabular}{l c ccc}
    \toprule
    \multirow{2}{*}{Rounds ($K$)} & \multirow{2}{*}{Storage (MB)} & \multicolumn{3}{c}{Wall-time Speedup} \\
    \cmidrule(lr){3-5}
    & & HumanEval & GSM8K & MT-Bench \\
    \midrule
    w/o Prior & \cellcolor{white}0.0 & 2.35$\times$ & 1.95$\times$ & 1.78$\times$ \\
    5         & \cellcolor{gray!15}4.7 & 2.72$\times$ & 2.21$\times$ & 1.88$\times$ \\
    \textbf{10}        & \cellcolor{cyan!10}\textbf{9.5} & \textbf{2.84}$\times$ & \textbf{2.30}$\times$ & \textbf{1.94}$\times$ \\
    15        & \cellcolor{pink!15}14.3& 2.86$\times$ & 2.31$\times$ & 1.94$\times$ \\
    20        & \cellcolor{pink!30}19.3& 2.86$\times$ & 2.33$\times$ & 1.96$\times$ \\
    \bottomrule
    \end{tabular}
    }
\vspace{-0.2in}    
\end{table}
\section{Conclusion}
We propose a parallel speculative decoding framework with a double-retrieval mechanism. \textsc{Double} utilizes iterative drafting to expand candidate lengths beyond theoretical limits and leverages target-side retrieval to repair potential rejections actively, thus compensating for the limitations of each model. Our approach achieves training-free and lossless speedup of $\textbf{5.3}\times$ even against EAGLE-3 and establishes a new SOTA for LLM inference. 

\section*{Limitations}
Our study is comprehensive, but has certain limitations that we plan to address in future research. In this study, we employ a unified datastore and synchronized retrieval depth for both draft and target models. While this configuration simplifies deployment and optimizes memory management, it may not fully exploit the distinct capabilities of each model. We believe these are minor issues and we will explore model-specific optimizations, such as adaptive retrieval depths and decoupled high-efficiency datastores, to further unlock the potential of parallel speculative decoding in future.

\section*{Acknowledgements}
This work is supported by National Science Foundation of China under grants 62576310, 62394341 and Zhejiang Provincial National Science Foundation of China under Grant No. LZ25F020007.

\section*{Ethics Statement}
The data and models utilized in this work are derived solely from publicly accessible resources with proper citations, so no sensitive information is involved. As \textsc{Double} is a speculative decoding framework designed to losslessly accelerate LLMs without parameter modification, it inherently inherits the biases and safety risks of the underlying models. It neither introduces new harmful capabilities nor mitigates existing ones. Therefore, standard safety guardrails and alignment techniques remain essential when deploying target models with \textsc{Double}.

\bibliography{custom}

@article{chen2024sequoia,
  title={Sequoia: Scalable and robust speculative decoding},
  author={Chen, Zhuoming and May, Avner and Svirschevski, Ruslan and Huang, Yu-Hsun and Ryabinin, Max and Jia, Zhihao and Chen, Beidi},
  journal={Advances in Neural Information Processing Systems},
  volume={37},
  pages={129531--129563},
  year={2024}
}

@article{team2023gemini,
  title={Gemini: a family of highly capable multimodal models},
  author={Team, Gemini and Anil, Rohan and Borgeaud, Sebastian and Alayrac, Jean-Baptiste and Yu, Jiahui and Soricut, Radu and Schalkwyk, Johan and Dai, Andrew M and Hauth, Anja and Millican, Katie and others},
  journal={arXiv preprint arXiv:2312.11805},
  year={2023}
}

@article{guo2025deepseek,
  title={Deepseek-r1: Incentivizing reasoning capability in llms via reinforcement learning},
  author={Guo, Daya and Yang, Dejian and Zhang, Haowei and Song, Junxiao and Zhang, Ruoyu and Xu, Runxin and Zhu, Qihao and Ma, Shirong and Wang, Peiyi and Bi, Xiao and others},
  journal={arXiv preprint arXiv:2501.12948},
  year={2025}
}

@article{agrawal2024adaedl,
  title={AdaEDL: Early Draft Stopping for Speculative Decoding of Large Language Models via an Entropy-based Lower Bound on Token Acceptance Probability},
  author={Agrawal, Sudhanshu and Jeon, Wonseok and Lee, Mingu},
  journal={arXiv preprint arXiv:2410.18351},
  year={2024}
}

@article{cai2024medusa,
  title={Medusa: Simple llm inference acceleration framework with multiple decoding heads},
  author={Cai, Tianle and Li, Yuhong and Geng, Zhengyang and Peng, Hongwu and Lee, Jason D and Chen, Deming and Dao, Tri},
  journal={arXiv preprint arXiv:2401.10774},
  year={2024}
}

@article{hinton2015distilling,
  title={Distilling the knowledge in a neural network},
  author={Hinton, Geoffrey and Vinyals, Oriol and Dean, Jeff},
  journal={arXiv preprint arXiv:1503.02531},
  year={2015}
}

@article{liu2024kangaroo,
  title={Kangaroo: Lossless self-speculative decoding for accelerating llms via double early exiting},
  author={Liu, Fangcheng and Tang, Yehui and Liu, Zhenhua and Ni, Yunsheng and Tang, Duyu and Han, Kai and Wang, Yunhe},
  journal={Advances in Neural Information Processing Systems},
  volume={37},
  pages={11946--11965},
  year={2024}
}

@article{li2024eagle,
  title={Eagle: Speculative sampling requires rethinking feature uncertainty},
  author={Li, Yuhui and Wei, Fangyun and Zhang, Chao and Zhang, Hongyang},
  journal={arXiv preprint arXiv:2401.15077},
  year={2024}
}

@article{du2024glide,
  title={Glide with a cape: A low-hassle method to accelerate speculative decoding},
  author={Du, Cunxiao and Jiang, Jing and Yuanchen, Xu and Wu, Jiawei and Yu, Sicheng and Li, Yongqi and Li, Shenggui and Xu, Kai and Nie, Liqiang and Tu, Zhaopeng and others},
  journal={arXiv preprint arXiv:2402.02082},
  year={2024}
}

@article{zhao2024ouroboros,
  title={Ouroboros: Generating Longer Drafts Phrase by Phrase for Faster Speculative Decoding},
  author={Zhao, Weilin and Huang, Yuxiang and Han, Xu and Xu, Wang and Xiao, Chaojun and Zhang, Xinrong and Fang, Yewei and Zhang, Kaihuo and Liu, Zhiyuan and Sun, Maosong},
  journal={arXiv preprint arXiv:2402.13720},
  year={2024}
}

@article{liu2024parallel,
  title={Parallel speculative decoding with adaptive draft length},
  author={Liu, Tianyu and Li, Yun and Lv, Qitan and Liu, Kai and Zhu, Jianchen and Hu, Winston},
  journal={arXiv preprint arXiv:2408.11850},
  year={2024}
}

@article{fu2024break,
  title={Break the sequential dependency of llm inference using lookahead decoding},
  author={Fu, Yichao and Bailis, Peter and Stoica, Ion and Zhang, Hao},
  journal={arXiv preprint arXiv:2402.02057},
  year={2024}
}

@article{chen2023accelerating,
  title={Accelerating large language model decoding with speculative sampling},
  author={Chen, Charlie and Borgeaud, Sebastian and Irving, Geoffrey and Lespiau, Jean-Baptiste and Sifre, Laurent and Jumper, John},
  journal={arXiv preprint arXiv:2302.01318},
  year={2023}
}

@article{sanh2019distilbert,
  title={DistilBERT, a distilled version of BERT: smaller, faster, cheaper and lighter},
  author={Sanh, Victor and Debut, Lysandre and Chaumond, Julien and Wolf, Thomas},
  journal={arXiv preprint arXiv:1910.01108},
  year={2019}
}

@inproceedings{frantar2023sparsegpt,
  title={Sparsegpt: Massive language models can be accurately pruned in one-shot},
  author={Frantar, Elias and Alistarh, Dan},
  booktitle={International Conference on Machine Learning},
  pages={10323--10337},
  year={2023},
  organization={PMLR}
}

@article{hu2022lora,
  title={Lora: Low-rank adaptation of large language models.},
  author={Hu, Edward J and Shen, Yelong and Wallis, Phillip and Allen-Zhu, Zeyuan and Li, Yuanzhi and Wang, Shean and Wang, Lu and Chen, Weizhu and others},
  journal={ICLR},
  volume={1},
  number={2},
  pages={3},
  year={2022}
}

@article{dao2022flashattention,
  title={Flashattention: Fast and memory-efficient exact attention with io-awareness},
  author={Dao, Tri and Fu, Dan and Ermon, Stefano and Rudra, Atri and R{\'e}, Christopher},
  journal={Advances in neural information processing systems},
  volume={35},
  pages={16344--16359},
  year={2022}
}

@inproceedings{leviathan2023fast,
  title={Fast inference from transformers via speculative decoding},
  author={Leviathan, Yaniv and Kalman, Matan and Matias, Yossi},
  booktitle={International Conference on Machine Learning},
  pages={19274--19286},
  year={2023},
  organization={PMLR}
}

@article{huang2024specdec++,
  title={Specdec++: Boosting speculative decoding via adaptive candidate lengths},
  author={Huang, Kaixuan and Guo, Xudong and Wang, Mengdi},
  journal={arXiv preprint arXiv:2405.19715},
  year={2024}
}

@inproceedings{Nallapati_Zhou_dos,   
title={Abstractive Text Summarization Using Sequence-to-Sequence RNNs and Beyond},  url={http://dx.doi.org/10.18653/v1/k16-1028},  DOI={10.18653/v1/k16-1028},  booktitle={Proceedings of The 20th SIGNLL Conference on Computational Natural Language Learning},  author={Nallapati, Ramesh and Zhou, Bowen and dos Santos, Cicero and Gulcehre, Caglar and Xiang, Bing},  year={2016},  month={Jan},  language={en-US}  }

@article{Cobbe_Kosaraju_Bavarian_Hilton_Nakano_Hesse_Schulman_2021,   
title={Training Verifiers to Solve Math Word Problems},  journal={Cornell University - arXiv,Cornell University - arXiv},  author={Cobbe, Karl and Kosaraju, Vineet and Bavarian, Mohammad and Hilton, Jacob and Nakano, Reiichiro and Hesse, Christopher and Schulman, John},  year={2021},  month={Oct},  language={en-US}  }

@article{chen2021codex,
  title={Evaluating Large Language Models Trained on Code},
  author={Mark Chen and Jerry Tworek and Heewoo Jun and Qiming Yuan and Henrique Ponde de Oliveira Pinto and Jared Kaplan and Harri Edwards and Yuri Burda and Nicholas Joseph and Greg Brockman et al.},
  year={2021},
  eprint={2107.03374},
  archivePrefix={arXiv},
  primaryClass={cs.LG}
}

@article{zheng2023judging,
  title={Judging LLM-as-a-judge with MT-Bench and Chatbot Arena},
  author={Lianmin Zheng and Wei-Lin Chiang and Ying Sheng and Siyuan Zhuang and Zhanghao Wu and Yonghao Zhuang and Zi Lin and Zhuohan Li and Dacheng Li and Eric.P Xing and Hao Zhang and JosephE. Gonzalez and Ion Stoica},
  year={2023},
  month={Jun},
  language={en-US}
}

@misc{deepseek-coder,
  author={Daya Guo and Qihao Zhu and Dejian Yang and Zhenda Xie and Kai Dong and Wentao Zhang and Guanting Chen and Xiao Bi and Y. Wu and Y.K. Li and Fuli Luo and Yingfei Xiong and Wenfeng Liang},
  title={DeepSeek-Coder: When the Large Language Model Meets Programming -- The Rise of Code Intelligence},
  journal={CoRR},
  volume={abs/2401.14196},
  year={2024},
  url={https://arxiv.org/abs/2401.14196}
}

@misc{grattafiori2024llama3herdmodels,
  title={The Llama 3 Herd of Models},
  author={Aaron Grattafiori and Abhimanyu Dubey and Abhinav Jauhri and Abhinav Pandey and Abhishek Kadian and Ahmad Al-Dahle and Aiesha Letman and Akhil Mathur and Alan Schelten and Alex Vaughan and Amy Yang and Angela Fan and Anirudh Goyal et al.},
  year={2024},
  eprint={2407.21783},
  archivePrefix={arXiv},
  primaryClass={cs.AI},
  url={https://arxiv.org/abs/2407.21783}
}

@article{pd_2024,   title={DistServe: Disaggregating Prefill and Decoding for Goodput-optimized Large Language Model Serving},  author={Zhong, Yinmin and Liu, Shengyu and Chen, Junda and Hu, Jianbo and Zhu, Yibo and Liu, Xuanzhe and Jin, Xin and Zhang, Hao},  year={2024},  month={Jan},  language={en-US}  }

@article{he2023rest,
  title={Rest: Retrieval-based speculative decoding},
  author={He, Zhenyu and Zhong, Zexuan and Cai, Tianle and Lee, Jason D and He, Di},
  journal={arXiv preprint arXiv:2311.08252},
  year={2023}
}

@article{xia2024swift,
  title={Swift: On-the-fly self-speculative decoding for llm inference acceleration},
  author={Xia, Heming and Li, Yongqi and Zhang, Jun and Du, Cunxiao and Li, Wenjie},
  journal={arXiv preprint arXiv:2410.06916},
  year={2024}
}

@article{timor2024distributed,
  title={Distributed speculative inference (dsi): Speculation parallelism for provably faster lossless language model inference},
  author={Timor, Nadav and Mamou, Jonathan and Korat, Daniel and Berchansky, Moshe and Pereg, Oren and Wasserblat, Moshe and Galanti, Tomer and Gordon, Michal and Harel, David},
  journal={arXiv preprint arXiv:2405.14105},
  year={2024}
}

@article{lu2025onpolicydistillation,
  author = {Kevin Lu and Thinking Machines Lab},
  title = {On-Policy Distillation},
  journal = {Thinking Machines Lab: Connectionism},
  year = {2025},
  note = {https://thinkingmachines.ai/blog/on-policy-distillation},
  doi = {10.64434/tml.20251026},
}

@article{li2025eagle,
  title={Eagle-3: Scaling up inference acceleration of large language models via training-time test},
  author={Li, Yuhui and Wei, Fangyun and Zhang, Chao and Zhang, Hongyang},
  journal={arXiv preprint arXiv:2503.01840},
  year={2025}
}

@inproceedings{zhang2024draft,
  title={Draft\& verify: Lossless large language model acceleration via self-speculative decoding},
  author={Zhang, Jun and Wang, Jue and Li, Huan and Shou, Lidan and Chen, Ke and Chen, Gang and Mehrotra, Sharad},
  booktitle={Proceedings of the 62nd Annual Meeting of the Association for Computational Linguistics (Volume 1: Long Papers)},
  pages={11263--11282},
  year={2024}
}

@article{choi2018pact,
  title={Pact: Parameterized clipping activation for quantized neural networks},
  author={Choi, Jungwook and Wang, Zhuo and Venkataramani, Swagath and Chuang, Pierce I-Jen and Srinivasan, Vijayalakshmi and Gopalakrishnan, Kailash},
  journal={arXiv preprint arXiv:1805.06085},
  year={2018}
}

@article{shen2025speculative,
  title={Speculative Decoding via Hybrid Drafting and Rollback-Aware Branch Parallelism},
  author={Shen, Yuhao and Shen, Junyi and Kong, Quan and Liu, Tianyu and Lu, Yao and Wang, Cong},
  journal={arXiv preprint arXiv:2506.01979},
  year={2025}
}

@article{brown2020language,
  title={Language models are few-shot learners},
  author={Brown, Tom and Mann, Benjamin and Ryder, Nick and Subbiah, Melanie and Kaplan, Jared D and Dhariwal, Prafulla and Neelakantan, Arvind and Shyam, Pranav and Sastry, Girish and Askell, Amanda and others},
  journal={Advances in neural information processing systems},
  volume={33},
  pages={1877--1901},
  year={2020}
}

@Book{MachineLearningI,
  editor = 	 "R. S. Michalski and J. G. Carbonell and T.
		  M. Mitchell",
  title = 	 "Machine Learning: An Artificial Intelligence
		  Approach, Vol. I",
  publisher = 	 "Tioga",
  year = 	 "1983",
  address =	 "Palo Alto, CA"
}

@misc{pld-saxena-2023,
    title = {Prompt Lookup Decoding},
    author = {Apoorv Saxena},
    year = {2023},
    month = {November},
    url = {https://github.com/apoorvumang/prompt-lookup-decoding/}
}

@inproceedings{rest-he-2024,
  title={REST: Retrieval-Based Speculative Decoding},
  author={He, Zhenyu and Zhong, Zexuan and Cai, Tianle and Lee, Jason and He, Di},
  booktitle={Proceedings of the 2024 Conference of the North American Chapter of the Association for Computational Linguistics: Human Language Technologies (Volume 1: Long Papers)},
  pages={1582--1595},
  year={2024}
}

@article{token-recycle-luo-2024,
  title={Turning Trash into Treasure: Accelerating Inference of Large Language Models with Token Recycling},
  author={Luo, Xianzhen and Wang, Yixuan and Zhu, Qingfu and Zhang, Zhiming and Zhang, Xuanyu and Yang, Qing and Xu, Dongliang and Che, Wanxiang},
  journal={arXiv preprint arXiv:2408.08696},
  year={2024}
}

@inproceedings{
RIA-zhang-2024,
title={Plug-and-Play: An Efficient Post-training Pruning Method for Large Language Models},
author={Yingtao Zhang and Haoli Bai and Haokun Lin and Jialin Zhao and Lu Hou and Carlo Vittorio Cannistraci},
booktitle={The Twelfth International Conference on Learning Representations},
year={2024},
url={https://openreview.net/forum?id=Tr0lPx9woF}
}

@misc{wanda-sun-2024,
      title={A Simple and Effective Pruning Approach for Large Language Models}, 
      author={Mingjie Sun and Zhuang Liu and Anna Bair and J. Zico Kolter},
      year={2024},
      eprint={2306.11695},
      archivePrefix={arXiv},
      primaryClass={cs.CL},
      url={https://arxiv.org/abs/2306.11695}, 
}

@misc{llm-streamline-chen-2024,
      title={Streamlining Redundant Layers to Compress Large Language Models}, 
      author={Xiaodong Chen and Yuxuan Hu and Jing Zhang and Yanling Wang and Cuiping Li and Hong Chen},
      year={2024},
      eprint={2403.19135},
      archivePrefix={arXiv},
      primaryClass={cs.CL},
      url={https://arxiv.org/abs/2403.19135}, 
}

@misc{shortgpt-men-2024,
      title={ShortGPT: Layers in Large Language Models are More Redundant Than You Expect}, 
      author={Xin Men and Mingyu Xu and Qingyu Zhang and Bingning Wang and Hongyu Lin and Yaojie Lu and Xianpei Han and Weipeng Chen},
      year={2024},
      eprint={2403.03853},
      archivePrefix={arXiv},
      primaryClass={cs.CL},
      url={https://arxiv.org/abs/2403.03853}, 
}

@article{awq-lin-2024,
  title={AWQ: Activation-aware Weight Quantization for On-Device LLM Compression and Acceleration},
  author={Lin, Ji and Tang, Jiaming and Tang, Haotian and Yang, Shang and Chen, Wei-Ming and Wang, Wei-Chen and Xiao, Guangxuan and Dang, Xingyu and Gan, Chuang and Han, Song},
  journal={Proceedings of Machine Learning and Systems},
  volume={6},
  pages={87--100},
  year={2024}
}

@inproceedings{smoothquant-xiao-2023,
  title={Smoothquant: Accurate and efficient post-training quantization for large language models},
  author={Xiao, Guangxuan and Lin, Ji and Seznec, Mickael and Wu, Hao and Demouth, Julien and Han, Song},
  booktitle={International Conference on Machine Learning},
  pages={38087--38099},
  year={2023},
  organization={PMLR}
}

@article{quarot-ashkboos-2024,
  title={Quarot: Outlier-free 4-bit inference in rotated llms},
  author={Ashkboos, Saleh and Mohtashami, Amirkeivan and Croci, Maximilian L and Li, Bo and Cameron, Pashmina and Jaggi, Martin and Alistarh, Dan and Hoefler, Torsten and Hensman, James},
  journal={arXiv preprint arXiv:2404.00456},
  year={2024}
}

@article{spinquant-liu-2024,
  title={SpinQuant--LLM quantization with learned rotations},
  author={Liu, Zechun and Zhao, Changsheng and Fedorov, Igor and Soran, Bilge and Choudhary, Dhruv and Krishnamoorthi, Raghuraman and Chandra, Vikas and Tian, Yuandong and Blankevoort, Tijmen},
  journal={arXiv preprint arXiv:2405.16406},
  year={2024}
}

@misc{compact-llm-muralidharan-2024,
      title={Compact Language Models via Pruning and Knowledge Distillation}, 
      author={Saurav Muralidharan and Sharath Turuvekere Sreenivas and Raviraj Joshi and Marcin Chochowski and Mostofa Patwary and Mohammad Shoeybi and Bryan Catanzaro and Jan Kautz and Pavlo Molchanov},
      year={2024},
      eprint={2407.14679},
      archivePrefix={arXiv},
      primaryClass={cs.CL},
      url={https://arxiv.org/abs/2407.14679}, 
}

@article{minitron-sreenivas-2024,
  title={Llm pruning and distillation in practice: The minitron approach},
  author={Sreenivas, Sharath Turuvekere and Muralidharan, Saurav and Joshi, Raviraj and Chochowski, Marcin and Patwary, Mostofa and Shoeybi, Mohammad and Catanzaro, Bryan and Kautz, Jan and Molchanov, Pavlo},
  journal={arXiv preprint arXiv:2408.11796},
  year={2024}
}

@misc{llama3-dubey-2024,
      title={The Llama 3 Herd of Models}, 
      author={Abhimanyu Dubey and Abhinav Jauhri and Abhinav Pandey and Abhishek Kadian and Ahmad Al-Dahle and Aiesha Letman et al.},
      year={2024},
      eprint={2407.21783},
      archivePrefix={arXiv},
      primaryClass={cs.AI},
      url={https://arxiv.org/abs/2407.21783}, 
}

@article{qwen2-yang-2024,
  title={Qwen2 technical report},
  author={Yang, An and Yang, Baosong and Hui, Binyuan and Zheng, Bo and Yu, Bowen and Zhou, Chang and Li, Chengpeng and Li, Chengyuan and Liu, Dayiheng and Huang, Fei and others},
  journal={arXiv preprint arXiv:2407.10671},
  year={2024}
}

@misc{gpt3-brown-2020,
      title={Language Models are Few-Shot Learners}, 
      author={Tom B. Brown and Benjamin Mann and Nick Ryder and Melanie Subbiah and Jared Kaplan and Prafulla Dhariwal and Arvind Neelakantan and Pranav Shyam and Girish Sastry and Amanda Askell and Sandhini Agarwal and Ariel Herbert-Voss and Gretchen Krueger and Tom Henighan and Rewon Child and Aditya Ramesh and Daniel M. Ziegler and Jeffrey Wu and Clemens Winter and Christopher Hesse and Mark Chen and Eric Sigler and Mateusz Litwin and Scott Gray and Benjamin Chess and Jack Clark and Christopher Berner and Sam McCandlish and Alec Radford and Ilya Sutskever and Dario Amodei},
      year={2020},
      eprint={2005.14165},
      archivePrefix={arXiv},
      primaryClass={cs.CL},
      url={https://arxiv.org/abs/2005.14165}, 
}

@article{spec-bench-xia-2024,
  title={Unlocking efficiency in large language model inference: A comprehensive survey of speculative decoding},
  author={Xia, Heming and Yang, Zhe and Dong, Qingxiu and Wang, Peiyi and Li, Yongqi and Ge, Tao and Liu, Tianyu and Li, Wenjie and Sui, Zhifang},
  journal={arXiv preprint arXiv:2401.07851},
  year={2024}
}

@article{mt-bench-zheng-2023,
  title={Judging llm-as-a-judge with mt-bench and chatbot arena},
  author={Zheng, Lianmin and Chiang, Wei-Lin and Sheng, Ying and Zhuang, Siyuan and Wu, Zhanghao and Zhuang, Yonghao and Lin, Zi and Li, Zhuohan and Li, Dacheng and Xing, Eric and others},
  journal={Advances in Neural Information Processing Systems},
  volume={36},
  pages={46595--46623},
  year={2023}
}

@article{liu2025logitspec,
  title={LogitSpec: Accelerating Retrieval-based Speculative Decoding via Next Next Token Speculation},
  author={Liu, Tianyu and Lv, Qitan and Li, Hao and Gao, Xing and Sun, Xiao},
  journal={arXiv preprint arXiv:2507.01449},
  year={2025}
}

@article{yang2025qwen3,
  title={Qwen3 technical report},
  author={Yang, An and Li, Anfeng and Yang, Baosong and Zhang, Beichen and Hui, Binyuan and Zheng, Bo and Yu, Bowen and Gao, Chang and Huang, Chengen and Lv, Chenxu and others},
  journal={arXiv preprint arXiv:2505.09388},
  year={2025}
}

@article{taori2023alpaca,
  title={Alpaca: A strong, replicable instruction-following model},
  author={Taori, Rohan and Gulrajani, Ishaan and Zhang, Tianyi and Dubois, Yann and Li, Xuechen and Guestrin, Carlos and Liang, Percy and Hashimoto, Tatsunori B},
  journal={Stanford Center for Research on Foundation Models. https://crfm. stanford. edu/2023/03/13/alpaca. html},
  volume={3},
  number={6},
  pages={7},
  year={2023}
}

@article{huang2025jakiro,
  title={Jakiro: Boosting speculative decoding with decoupled multi-head via moe},
  author={Huang, Haiduo and Yang, Fuwei and Liu, Zhenhua and Xu, Yixing and Li, Jinze and Liu, Yang and Yin, Xuanwu and Li, Dong and Ren, Pengju and Barsoum, Emad},
  journal={arXiv preprint arXiv:2502.06282},
  year={2025}
}

@article{yang2025longspec,
  title={LongSpec: Long-Context Lossless Speculative Decoding with Efficient Drafting and Verification},
  author={Yang, Penghui and Du, Cunxiao and Zhang, Fengzhuo and Wang, Haonan and Pang, Tianyu and Du, Chao and An, Bo},
  journal={arXiv preprint arXiv:2502.17421},
  year={2025}
}

@misc{ji2026foresttreeslooselyspeculative,
      title={See the Forest for the Trees: Loosely Speculative Decoding via Visual-Semantic Guidance for Efficient Inference of Video LLMs}, 
      author={Yicheng Ji and Jun Zhang and Jinpeng Chen and Cong Wang and Lidan Shou and Gang Chen and Huan Li},
      year={2026},
      eprint={2604.05650},
      archivePrefix={arXiv},
      primaryClass={cs.CL},
      url={https://arxiv.org/abs/2604.05650}, 
}

@misc{zeng2026hybridkvhybridkvcache,
      title={HybridKV: Hybrid KV Cache Compression for Efficient Multimodal Large Language Model Inference}, 
      author={Bowen Zeng and Feiyang Ren and Jun Zhang and Xiaoling Gu and Ke Chen and Lidan Shou and Huan Li},
      year={2026},
      eprint={2604.05887},
      archivePrefix={arXiv},
      primaryClass={cs.AI},
      url={https://arxiv.org/abs/2604.05887}, 
}

@misc{zhang2026efficientinferencelargevisionlanguage,
      title={Efficient Inference for Large Vision-Language Models: Bottlenecks, Techniques, and Prospects}, 
      author={Jun Zhang and Yicheng Ji and Feiyang Ren and Yihang Li and Bowen Zeng and Zonghao Chen and Ke Chen and Lidan Shou and Gang Chen and Huan Li},
      year={2026},
      eprint={2604.05546},
      archivePrefix={arXiv},
      primaryClass={cs.CL},
      url={https://arxiv.org/abs/2604.05546}, 
}

@article{wang2026fbs,
  title={FBS: Modeling Native Parallel Reading inside a Transformer},
  author={Wang, Tongxi},
  journal={arXiv preprint arXiv:2601.21708},
  year={2026}
}

@article{wang2026perm,
  title={PERM: Psychology-grounded Empathetic Reward Modeling for Large Language Models},
  author={Wang, Chengbing and Zheng, Wuqiang and Zhang, Yang and Zhu, Fengbin and Cheng, Junyi and Xie, Yi and Wang, Wenjie and Feng, Fuli},
  journal={arXiv preprint arXiv:2601.10532},
  year={2026}
}

@article{zhang2026logical,
  title={Logical Phase Transitions: Understanding Collapse in LLM Logical Reasoning},
  author={Zhang, Xinglang and Zhang, Yunyao and Chen, ZeLiang and Yu, Junqing and Yang, Wei and Song, Zikai},
  journal={arXiv preprint arXiv:2601.02902},
  year={2026}
}

@article{zhang2025ambiguity,
  title={From ambiguity to verdict: A Semiotic-Grounded Multi-Perspective agent for LLM logical reasoning},
  author={Zhang, Yunyao and Zhang, Xinglang and Sheng, Junxi and Li, Wenbing and Yu, Junqing and Chen, Yi-Ping Phoebe and Yang, Wei and Song, Zikai},
  journal={arXiv preprint arXiv:2509.24765},
  year={2025}
}

@inproceedings{shen2025hetero,
  title={Hetero 2 Pipe: Pipelining Multi-DNN Inference on Heterogeneous Mobile Processors under Co-Execution Slowdown},
  author={Shen, Yuhao and Wang, Zichen and Wang, Tianyi and Gu, Chaojie and Wen, Zhenyu and Shu, Yuanchao and Wang, Cong},
  booktitle={2025 IEEE 45th International Conference on Distributed Computing Systems (ICDCS)},
  pages={483--493},
  year={2025},
  organization={IEEE}
}

@article{kong2026parallelvlm,
  title={ParallelVLM: Lossless Video-LLM Acceleration with Visual Alignment Aware Parallel Speculative Decoding},
  author={Kong, Quan and Shen, Yuhao and Ji, Yicheng and Li, Huan and Wang, Cong},
  journal={arXiv preprint arXiv:2603.19610},
  year={2026}
}

@article{kong2026vision,
  title={Vision-TTT: Efficient and Expressive Visual Representation Learning with Test-Time Training},
  author={Kong, Quan and Xiao, Yanru and Shen, Yuhao and Wang, Cong},
  journal={arXiv preprint arXiv:2603.00518},
  year={2026}
}

@article{wu2026atlas,
  title={Atlas: Orchestrating Heterogeneous Models and Tools for Multi-Domain Complex Reasoning},
  author={Wu, Jinyang and Zhai, Guocheng and Jin, Ruihan and Yuan, Jiahao and Shen, Yuhao and Zhang, Shuai and Wen, Zhengqi and Tao, Jianhua},
  journal={arXiv preprint arXiv:2601.03872},
  year={2026}
}

@article{wu2026ssl,
  title={SSL: Sweet Spot Learning for Differentiated Guidance in Agentic Optimization},
  author={Wu, Jinyang and Yang, Changpeng and Shen, Yuhao and Xu, Fangzhi and Ni, Bolin and Liao, Chonghua and Liu, Yuchen and Wang, Hongzhen and Nie, Shuai and Zhang, Shuai and others},
  journal={arXiv preprint arXiv:2601.22491},
  year={2026}
}

@article{wu2026spark,
  title={Spark: Strategic Policy-Aware Exploration via Dynamic Branching for Long-Horizon Agentic Learning},
  author={Wu, Jinyang and Yang, Shuo and Yang, Changpeng and Shen, Yuhao and Zhang, Shuai and Wen, Zhengqi and Tao, Jianhua},
  journal={arXiv preprint arXiv:2601.20209},
  year={2026}
}

@article{liu2026talon,
  title={TALON: Confidence-Aware Speculative Decoding with Adaptive Token Trees},
  author={Liu, Tianyu and Lv, Qitan and Shen, Yuhao and Sun, Xiao and Sun, Xiaoyan},
  journal={arXiv preprint arXiv:2601.07353},
  year={2026}
}

@inproceedings{sun2025causalabstain,
  title={Causalabstain: Enhancing multilingual llms with causal reasoning for trustworthy abstention},
  author={Sun, Yuxi and Zuo, Aoqi and Gao, Wei and Ma, Jing},
  booktitle={Findings of the Association for Computational Linguistics: ACL 2025},
  pages={14060--14076},
  year={2025}
}

@misc{sun2026factecausalityinspiredevaluationtrustworthy,
      title={FACT-E: Causality-Inspired Evaluation for Trustworthy Chain-of-Thought Reasoning}, 
      author={Yuxi Sun and Aoqi Zuo and Haotian Xie and Wei Gao and Mingming Gong and Jing Ma},
      year={2026},
      eprint={2604.10693},
      archivePrefix={arXiv},
      primaryClass={cs.AI},
      url={https://arxiv.org/abs/2604.10693}, 
}

@misc{wang2025fakesvvlmtamingvlmdetecting,
      title={FakeSV-VLM: Taming VLM for Detecting Fake Short-Video News via Progressive Mixture-Of-Experts Adapter}, 
      author={Junxi Wang and Yaxiong Wang and Lechao Cheng and Zhun Zhong},
      year={2025},
      eprint={2508.19639},
      archivePrefix={arXiv},
      primaryClass={cs.MM},
      url={https://arxiv.org/abs/2508.19639}, 
}

@misc{wang2026streammecolongtermagentmemory,
      title={StreamMeCo: Long-Term Agent Memory Compression for Efficient Streaming Video Understanding}, 
      author={Junxi Wang and Te Sun and Jiayi Zhu and Junxian Li and Haowen Xu and Zichen Wen and Xuming Hu and Zhiyu Li and Linfeng Zhang},
      year={2026},
      eprint={2604.09000},
      archivePrefix={arXiv},
      primaryClass={cs.CV},
      url={https://arxiv.org/abs/2604.09000}, 
}

@article{wu2024beyond,
  title={Beyond examples: High-level automated reasoning paradigm in in-context learning via mcts},
  author={Wu, Jinyang and Feng, Mingkuan and Zhang, Shuai and Che, Feihu and Wen, Zengqi and Liao, Chonghua and Tao, Jianhua},
  journal={arXiv preprint arXiv:2411.18478},
  year={2024}
}

@article{shen2025batch,
  title={Batch Query Processing and Optimization for Agentic Workflows},
  author={Shen, Junyi and Wadlom, Noppanat and Lu, Yao},
  journal={arXiv preprint arXiv:2509.02121},
  year={2025}
}

@article{shen2025flowmesh,
  title={FlowMesh: A Service Fabric for Composable LLM Workflows},
  author={Shen, Junyi and Wadlom, Noppanat and Zhou, Lingfeng and Wang, Dequan and Miao, Xu and Fang, Lei and Lu, Yao},
  journal={arXiv preprint arXiv:2510.26913},
  year={2025}
}

@misc{an2026flowhijackdynamicsawarebackdoorattack,
      title={FlowHijack: A Dynamics-Aware Backdoor Attack on Flow-Matching Vision-Language-Action Models}, 
      author={Xinyuan An and Tao Luo and Gengyun Peng and Yaobing Wang and Kui Ren and Dongxia Wang},
      year={2026},
      eprint={2604.09651},
      archivePrefix={arXiv},
      primaryClass={cs.CV},
      url={https://arxiv.org/abs/2604.09651}, 
}

\clearpage
\appendix

\setcounter{tocdepth}{-1}
\addtocontents{toc}{\protect\setcounter{tocdepth}{2}}
\tableofcontents

\section{Procedures of \textsc{Double}}
The detailed procedures of \textsc{Double} are described in Algorithm~\ref{alg:double_decoding_part1} and Algorithm~\ref{alg:double_decoding_part2}.
\begin{algorithm}[t]
    \caption{Double Retrieval Speculative Parallelism (\textsc{Double}) - Part I.}\label{alg:double_decoding_part1}
    \renewcommand{\algorithmicrequire}{\textbf{Input:}}
    \definecolor{desc}{RGB}{128,128,128}
    \definecolor{acc}{RGB}{34,139,34}
    \definecolor{rej}{RGB}{178,34,34}
    
    \begin{algorithmic}[1]
        \REQUIRE{Draft model $M_q$, Target model $M_p$, Datastore $\mathcal{D}$, Prefix $\mathbf{x}$, Gamma $\gamma$.}
        \STATE \textcolor{desc}{$\triangleright$ Initialization}
        \STATE mode $\leftarrow$ ``pre-verify'', $prev\_tokens \leftarrow \gamma$
        
        \WHILE{not EndOfSequence}
            \STATE $L \leftarrow |\mathbf{x}|$
            
            \STATE \textcolor{desc}{$\triangleright$ \textbf{Parallel Generation}}
            \STATE  $X_q, n_q \leftarrow \textsc{Retrieval}(M_q, \mathbf{x}, \mathcal{D})$
            \STATE $X_p, n_p \leftarrow \textsc{Retrieval}(M_p, \mathbf{x}, \mathcal{D})$
            \STATE Gather probabilities: 
            \STATE $\text{Prob}_q[L-prev\_tokens-1:L]$, 
            \STATE $\text{Prob}_p[L-prev\_tokens-1:L]$.
            
            \STATE \textcolor{desc}{$\triangleright$ \textbf{Pre-verify Mode}}
            \IF{mode = ``pre-verify''}
                \STATE Find first reject position $n$ in $X_q[-n_q:]$ using $\text{Prob}_p, \text{Prob}_q$
                \IF{$n = n_p$ (all accepted)}
                    \STATE \textcolor{acc}{$\checkmark$ $\mathbf{x} \leftarrow \mathbf{x} + X_q[-n_q:]$, mode $\leftarrow$ ``post-verify''}
                    \STATE $prev\_tokens \leftarrow n_q$
                \ELSE
                    \STATE \textcolor{rej}{$\times$ $\mathbf{x} \leftarrow \mathbf{x} + X_p[:n]$, rollback $M_q$}
                    \STATE $prev\_tokens \leftarrow n_p$
                \ENDIF
            \ENDIF
            
            \STATE \textcolor{desc}{Continued in Algorithm~\ref{alg:double_decoding_part2}}
        \ENDWHILE
    \end{algorithmic}
\end{algorithm}
\vspace*{-0.1in}

\section{Detailed Theoretical Analysis}
\label{sec:appendix_theoretical}

In this section, we provide the formal derivation and proofs for the theorems presented in Section~\ref{sec:speed ceiling}.

\subsection{Preliminaries and Notations}
To be consistent with the main text, let $\alpha \in [0, 1]$ denote the token acceptance rate and $\gamma$ the draft length. Let $T_p$ and $T_q$ represent the single-token inference latency for the target and draft models, respectively. The speed ratio is defined as $C = T_p / T_q$.

Following standard modeling~\citep{chen2023accelerating, liu2024parallel}, we observe the accepted token distribution is bimodal~\citep{shen2025speculative} and model the decoding process as a sequence of $k-1$ fully accepted rounds followed by a single rejected round in the $k$-th iteration. Here, $k$ represents the expected number of rounds until a rejection occurs:
\begin{equation}
\small
    k = \mathbb{E}[\text{rounds}] = \frac{1}{1 - \alpha^\gamma}
\end{equation}

\subsection{Proof of Theorem 2 (Multi-Round)}
\textbf{Theorem 2}. \textit{The expected number of tokens generated over $k$ rounds is},
\begin{equation}
\small
\mathbb{E}[L_k] = (k-1)(\gamma+1) + \frac{1 - \alpha^{\gamma+1}}{1 - \alpha}.    
\end{equation}

\paragraph{Proof}.
In standard SD, verification is sequential. The total expected length $\mathbb{E}[L_k]$ consists of $k-1$ full rounds (each contributing $\gamma+1$ tokens: $\gamma$ accepted draft tokens + 1 verification token) and one final truncated round.
The expected length of the final truncated round $\mathbb{E}[L_{\text{last}}]$ follows a geometric sum based on the acceptance rate $\alpha$:
\begin{equation}
\small
\begin{aligned}
    \mathbb{E}[L_k] &= (k-1)(\gamma+1) + \mathbb{E}[L_{\text{last}}] \\
    &= (k-1)(\gamma+1) + \sum_{i=0}^{\gamma} \alpha^i \\
    &= (k-1)(\gamma+1) + \frac{1 - \alpha^{\gamma+1}}{1 - \alpha}
\end{aligned}
\end{equation}
Assuming verification time dominates (Under compute-bound, $T_{\text{AR}} \approx T_p$), the speedup $S_{\text{SD}}$ is:
\begin{equation}
\small
\label{eq:ssd_def}
\begin{aligned}
    S_{\text{SD}} &= \frac{\mathbb{E}[L_k] \cdot T_{\text{AR}}}{k (\gamma T_q + T_p)} \\
                  &= \frac{\mathbb{E}[L_k] \cdot C}{k(\gamma + C)}
\end{aligned}
\end{equation}

\subsection{Proof of Theorem 3 (Speedup Ceiling)}
\textbf{Theorem 3}. \textit{The theoretical speedup $S_{\text{PSD}}$ strictly dominates $S_{\text{SD}}$ but is upper bounded by $C$: $S_{\text{SD}} \le S_{\text{PSD}} \le C$.}

\paragraph{1. Formulation of PSD Speedup}
Parallel Speculative Decoding (PSD) pipelines the draft and verification phases. The expected accepted length $\mathbb{E}[L_p]$ in PSD differs from SD because the verification token of the previous round is not "free" for drafting. Based on the relationship derived in~\citet{liu2024parallel}:
\begin{equation}
\small
    \mathbb{E}[L_p] = \mathbb{E}[L_k] - (k-1)
\end{equation}
Assume optimal pipelining where the cost per round is dominated by $\max(\gamma T_q, T_p)$ (typically $\gamma T_q \approx T_p$, i.e., $\gamma \approx C$), the speedup $S_{\text{PSD}}$ is:
\begin{equation}
\small
\label{eq:spd_def}
\begin{aligned}
    S_{\text{PSD}} &= \frac{\mathbb{E}[L_p] \cdot C}{k \gamma + C} \\
                   &= \frac{(\mathbb{E}[L_k] - k + 1) \cdot C}{k \gamma + C}
\end{aligned}
\end{equation}

\paragraph{2. Proof of $S_{\text{PSD}} \ge S_{\text{SD}}$}
Let $N_{\text{SD}} = \mathbb{E}[L_k] \cdot C$ and $D_{\text{SD}} = k(\gamma + C)$ be the numerator and denominator of $S_{\text{SD}}$.
Define a shift term $\Delta = (k-1)C$. Since $k \ge 1, C > 0$, we have $\Delta \ge 0$.
Expressing $S_{\text{PSD}}$ in terms of $S_{\text{SD}}$ components:
\begin{align}
\small
    N_{\text{PSD}} &= (\mathbb{E}[L_k] - k + 1) C = N_{\text{SD}} - \Delta \\
    D_{\text{PSD}} &= k\gamma + C = D_{\text{SD}} - \Delta
\end{align}
We analyze the function $f(x) = \frac{N_{\text{SD}} - x}{D_{\text{SD}} - x}$. Its derivative is $f'(x) = \frac{N_{\text{SD}} - D_{\text{SD}}}{(D_{\text{SD}}-x)^2}$.
Given a valid speedup $S_{\text{SD}} \ge 1 \implies N_{\text{SD}} \ge D_{\text{SD}}$, thus $f'(x) \ge 0$.
Since $\Delta \ge 0$, it follows that $f(\Delta) \ge f(0)$, implying:
\begin{equation}
\small
    S_{\text{PSD}} = \frac{N_{\text{SD}} - \Delta}{D_{\text{SD}} - \Delta} \ge \frac{N_{\text{SD}}}{D_{\text{SD}}} = S_{\text{SD}}
\end{equation}

\paragraph{3. Proof of Upper Bound $C$}
We examine the upper bound of $S_{\text{PSD}}$ from Eq.~\eqref{eq:spd_def}. In the ideal scenario (perfect acceptance, $\alpha=1$), the maximum length generated per round is limited to $\gamma$. Thus, $\mathbb{E}[L_p] \le k\gamma$. Substituting this:
\begin{equation}
\small
\begin{aligned}
    S_{\text{PSD}} &\le \frac{k\gamma \cdot C}{k\gamma + C} \\
                   &= C \cdot \left( \frac{k\gamma}{k\gamma + C} \right)
\end{aligned}
\end{equation}
Since $k\gamma > 0$ and $C > 0$, the term $\frac{k\gamma}{k\gamma + C}$ is strictly less than 1. Therefore, $S_{\text{PSD}} < C$. Combining paragraphs 2 and 3, we conclude:
\begin{equation}
\small
    S_{\text{SD}} \le S_{\text{PSD}} \le C.
\end{equation}
\hfill 

\begin{algorithm}[t]
    \caption{Double Retrieval Speculative Parallelism (\textsc{Double}) - Part II.}\label{alg:double_decoding_part2}
    \definecolor{desc}{RGB}{128,128,128}
    \definecolor{acc}{RGB}{34,139,34}
    \definecolor{rej}{RGB}{178,34,34}
    
    \begin{algorithmic}[1]
        \STATE \textcolor{desc}{$\triangleright$ \textbf{Post-verify Mode}}
        \IF{mode = ``post-verify''}
            \STATE Find first reject $n$ in $X_q[0:prev\_tokens-1]$ using $\text{Prob}_p, \text{Prob}_q$
            
            \IF{$n = prev\_tokens - 1$}
                \STATE \textcolor{desc}{$\triangleright$ Previous verified, check new tokens}
                \IF{$n_p \geq n_q$}
                    \STATE \textcolor{acc}{$\checkmark$ $\mathbf{x} \leftarrow \mathbf{x} + X_p$, mode $\leftarrow$ ``pre-verify''}
                \ELSE
                    \STATE Find reject $n$ in $X_q[prev\_tokens-1:prev\_tokens+n_p-1]$
                    \IF{all verified}
                        \STATE \textcolor{acc}{$\checkmark$ $\mathbf{x} \leftarrow \mathbf{x} + X_q[-n_q:]$, $prev\_tokens \leftarrow n_q$}
                    \ELSE
                        \STATE \textcolor{rej}{$\times$ $\mathbf{x} \leftarrow \mathbf{x} + X_p$, mode $\leftarrow$ "pre-verify"}
                    \ENDIF
                \ENDIF
            \ELSE
                \STATE \textcolor{rej}{$\times$ Reject at $n$: sample $t \sim \max(\text{Prob}_p[n] - \text{Prob}_q[n], 0)$}
                \STATE $\mathbf{x} \leftarrow \mathbf{x}[:L-prev\_tokens+n+1] + t$
                \STATE mode $\leftarrow$ ``pre-verify'', rollback both models
            \ENDIF
        \ENDIF
        
        \STATE $\mathcal{D} \leftarrow \text{Update}(\mathcal{D}, \mathbf{x})$
    \end{algorithmic}
\end{algorithm}

\vspace*{-0.1in}
\section{Evaluation Details} \label{sec:appendix_eval}
For reproducibility, we discuss the experimental setup (Section~\ref{sec:experiments}) in detail and the source code of this project will be made available at a later time.

\subsection{Experimental Setting}
\label{sec:exp_setting}

\paragraph{Tasks and Datasets}
We evaluate \textsc{Double} across a diverse suite of LLM configurations, ranging from lightweight to large-scale architectures: LLaMA-2 (7B/70B)~\citep{grattafiori2024LLaMA3herdmodels}, LLaMA-3 (8B/70B)~\citep{llama3-dubey-2024}, Deepseek-Coder (1.3B/33B)~\citep{deepseek-coder}, and Qwen3 (0.6B/14B/32B)~\citep{yang2025qwen3}. To ensure comprehensive coverage, our benchmarks span code generation, mathematical reasoning, summarization, and general instruction following: HumanEval~\citep{chen2021codex}, GSM8K~\citep{Cobbe_Kosaraju_Bavarian_Hilton_Nakano_Hesse_Schulman_2021}, CNN/DM~\citep{Nallapati_Zhou_dos}, Alpaca~\citep{taori2023alpaca}, and MT-Bench~\citep{mt-bench-zheng-2023} following the set of EAGLE-3.

\paragraph{Baselines}
To validate the effectiveness of \textsc{Double}, we compare it against a comprehensive set of representative methods categorized by their acceleration mechanism:
\begin{itemize}[leftmargin=*, itemsep=0pt, topsep=2pt]
    \item \textbf{Standard SD}~\citep{chen2023accelerating}: The canonical \textit{draft-then-verify} framework. We utilize the same draft models as \textsc{Double} but execute them sequentially without retrieval augmentation.
    \item \textbf{Target-side Retrieval}: We include methods that accelerate inference solely using the target model. \textbf{Lookahead}~\citep{fu2024break} employs Jacobi iteration for multi-branch generation without a draft model. \textbf{PLD}~\citep{pld-saxena-2023} matches the current prefix against the prompt to reuse computations. \textbf{Token Recycling}~\citep{token-recycle-luo-2024} leverages past key-value pairs to predict future tokens.
    \item \textbf{Draft-side Retrieval}: We compare against \textbf{Ouroboros}~\citep{zhao2024ouroboros}, which constructs an $n$-gram index from the draft model's generation history to propose candidate suffixes, aiming to extend the draft length.
    \item \textbf{Parallel SD}: We use \textbf{PEARL}~\citep{liu2024parallel} as the primary parallel baseline. PEARL pipelines the drafting and verification phases to hide draft latency but relies on standard autoregressive drafting without retrieval guidance.
    \item \textbf{Training-based Methods}: We include \textbf{EAGLE-3}~\citep{li2025eagle}, the current state-of-the-art method that utilizes a lightweight training layer to incorporate feature-level history for more accurate drafting.
\end{itemize}
All baselines are reproduced from their official codebases using the optimal configurations. Experiments are conducted under identical conditions to avoid implementation bias. 

\paragraph{Implementation and Metrics}
All experiments are conducted on 8 NVIDIA A100 (80GB) GPUs. Model parallelism is applied where necessary: models under 33B parameters run on a single device, and 70B models are distributed across TWO devices. We use greedy sampling with a batch size of 1 for the main experiments. The draft length is dynamically set to $\gamma = \lceil C \rceil$, where $C$ represents the average speed ratio $T_p/T_q$ across datasets in off-the-shelf framework. \textbf{For EAGLE-3}, it utilizes a lightweight training layer to reduce the draft model cost and the $C$ is different.
We set the default depth $d=10$, prior round $K=10$, and n-gram $n=3$.

We report three key metrics: \textbf{Wall-Time Speedup}, and \textbf{Mean Accepted Length ($M$)}. In the context of parallel frameworks (PEARL and \textsc{Double}), $M$ denotes the \textit{continuously accepted length}~\citep{liu2024parallel} achieved through multiple rounds of pipelined generation, rather than the single-round acceptance of vanilla SD. Additionally, we analyze the Average Matched Tokens (\text{AMT}) defined as the expectation of $s$, which directly determines the effective speedup, serving as a proxy for the precision-efficiency trade-off.

\begin{table}[h]
\centering
\resizebox{1\columnwidth}{!}{  
\begin{tabular}{ccccc}
\toprule
    \textbf{Models} & \textbf{Layers} & \textbf{dim} & \textbf{FFN dim} & \textbf{Vocabulary size} \\
\midrule
Deepseek 1.3B & 24  & 2048  & 5504  & 32256   \\
Deepseek 33B & 62 & 7168  & 19200  & 32256   \\
LLaMA-2 7B & 32 & 4096 & 11008  & 32000  \\
LLaMA-2 70B & 80 & 8192  & 28672  & 32000 \\
LLaMA-3.1 8B & 32 & 4096 & 14336  & 128256   \\
LLaMA-3.3 70B & 80 & 8192  & 28672  & 128256 \\
Qwen-3 0.6B & 28 & 1024  & 3072  & 151936 \\
Qwen-3 14B & 40 & 5120  & 17408  & 151936 \\
Qwen-3 32B & 64 & 5120  & 25600 & 151936 \\
\bottomrule
\end{tabular}
}
\vspace*{-0.1in}
\caption{Model configurations.}
\label{tab:model_conf}
\vspace*{-0.1in}
\end{table}

 \section{More Experimental Results and Discussions} \label{sec:appendix_results}

\subsection{Scalability to High Batch Sizes}
\label{sec:appendix_batch}

Our primary evaluation focuses on latency-critical scenarios ($bs=1$), which represent the standard use case for real-time interaction. However, in high-throughput settings with larger batch sizes, the GPU's compute-bound nature may saturate, and the overhead of maintaining diverse dynamic tree structures for each request can become non-trivial.

To assess scalability, we leverage \textsc{Nano-Pearl}~\citep{liu2024parallel}, a novel Parallel SD framework built upon Nano-vLLM. It features a high-concurrency engine optimized with custom kernels for industrial acceleration. We benchmark \textsc{Nano-Pearl} against EAGLE-3 on Qwen3-32B across batch sizes $bs \in \{1, \dots, 16\}$. As detailed in Table~\ref{tab:high_batch}, EAGLE-3 suffers significant degradation at $bs=16$ (dropping to $1.13\times$). This decline is primarily attributed to the memory access overhead and compute-bound caused by the complex, non-contiguous tree attention masks. 

In contrast, \textsc{Nano-Pearl} unlocks high-throughput potential through chained drafting and parallel decoding, sustaining a robust speedup of $1.73\times$ even at $bs=16$. Similarly, \textsc{Double} is inherently suited for this architecture; its linear retrieval structure eliminates the overhead of dynamic masking. By integrating with \textsc{Nano-Pearl} and dynamically tuning retrieval depth, \textsc{Double} can further enhance speedups in high-throughput environments. We are currently integrating \textsc{Double} into \textsc{Nano-Pearl} and plan to release it as a feature branch to the community, contributing to robust industrial deployment. Future work will also explore migrating this parallel SD paradigm to comprehensive inference engines such as vLLM and SGLang.

\begin{table}[h]
    \centering
    \caption{High-Batch performance comparison on Qwen3-32B. While EAGLE-3 degrades significantly due to mask overheads, \textsc{Nano-Pearl} maintains robust acceleration under high concurrency ($bs=16$).}
    \label{tab:high_batch}
    \resizebox{0.48\textwidth}{!}{%
    \small 
    \begin{tabular}{lccccc}
    \toprule
    & \multicolumn{5}{c}{Batch Size (HumanEval)} \\
    \cmidrule(lr){2-6}
    & 1 & 2 & 4 & 8 & 16 \\
    \toprule
    EAGLE-3 & 1.98$\times$ & 1.83$\times$ & 1.67$\times$ & 1.34$\times$ & 1.13$\times$\\
    \cellcolor{pink!30}\textbf{\textsc{Nano-Pearl}}  & \cellcolor{pink!30}\textbf{2.64$\times$} & \cellcolor{pink!30}\textbf{2.33$\times$} & \cellcolor{pink!30}\textbf{2.17$\times$} & \cellcolor{pink!30}\textbf{1.94$\times$} & \cellcolor{pink!30}\textbf{1.73$\times$}\\
    \bottomrule
    \end{tabular}
    }
\vspace*{-0.1in}
\end{table}

\subsection{Proof of Lossless Acceleration}
\label{sec:appendix_lossless}

In this section, we provide both theoretical and empirical verification to demonstrate the lossless nature of \textsc{Double}. \textbf{We show that our Target-Guided Verification mechanism strictly preserves the target model's output distribution $p(x)$ under both greedy ($T=0$) and stochastic ($T>0$) decoding regimes.}

\subsubsection{Theoretical Guarantee}
The fundamental premise of Speculative Decoding (SD) and Retrieval-based SD is universally acknowledged as \textbf{lossless}: the final output distribution must rigorously match the target distribution $p(x)$, independent of the draft source. Our framework strictly adheres to this standard verification protocol~\citep{leviathan2023fast, chen2023accelerating}. We analyze the validity of our mechanism under two distinct settings:

\paragraph{Greedy Decoding ($T=0$)}
In the greedy setting, the target distribution degenerates into a Dirac delta function: $p(x) = \delta(x - \arg\max p(\cdot))$. We provide the formal proof for the \textit{Correction} and \textit{Extension} phases:

\begin{itemize}[leftmargin=*]
    \item \textbf{Correction Mechanism ($y_i$):} Standard rejection sampling requires drawing a correction from the residual distribution when a draft token $x_i$ diverges from the target. In greedy decoding, the target distribution $p(x)$ is a Dirac delta function centered at the top-1 prediction $y_i$. This forces the residual sampling to deterministically select $y_i$. Therefore, replacing the rejected $x_i$ with $y_i$ is mathematically equivalent to the formal rejection sampling process.
    
    \item \textbf{Validity of Extension ($y_{i+1} \dots y_{s+1}$):} We explicitly address the concern regarding whether appending tokens from $Y_p$ strictly preserves the target distribution. It is crucial to clarify that $Y_p$ is \textbf{not} a heuristic sequence retrieved directly from the datastore. As defined in Sec.~\ref{sec:prelim2} (Eq.~2), $Y_p$ represents the \textbf{verified output} of the target model $\mathcal{M}_p$ computed during the parallel verification step.
    
    Specifically, $\mathcal{M}_p$ employs a causal mask to process the retrieved candidates in parallel. This means $\mathcal{M}_p$ calculates the conditional probability $p(y_k \mid \text{prefix}, y_1, \dots, y_{k-1})$ for all positions $k$ simultaneously within the same forward pass. Consequently, if $y_i$ is selected as the correction, the subsequent token $y_{i+1}$ has already been computed by $\mathcal{M}_p$ conditioned specifically on the path including $y_i$. Thus, the sequence $\{y_{i+1}, \dots, y_{s+1}\}$ constitutes the ground-truth greedy path of the target model. Utilizing these pre-calculated tokens is not an approximation but a direct application of the target model's own generation, ensuring strict losslessness without redundant re-computation.
\end{itemize}

According to the standard rejection sampling criterion~\citep{leviathan2023fast}, if a draft token $x_i$ diverges from the target (i.e., $x_i \neq \arg\max p(\cdot)$), a correction must be sampled from the residual distribution,
\begin{equation}
\small
x'_{i} \sim \text{norm}(\max(0, p(x) - q(x)))
\end{equation}
Under greedy sampling, since $p(x)$ has a probability mass of 1 ($q(x)=0$) at the optimal token $y_i$, the sampled correction $x'_i$ collapses deterministically to $y_i$. Consequently, replacing the rejected draft token $x_i$ with the target's retrieved token $y_i$ is mathematically equivalent to performing rejection sampling. Furthermore, because the corrected prefix aligns with the target model's deterministic path, the subsequent tokens in the retrieved chain ($y_{i+1}, \dots$) remain valid. This allows \textsc{Double} to utilize \textbf{Forward Guidance} (Extension) losslessly in greedy scenarios.

\paragraph{Stochastic Sampling ($T > 0$)}
Under non-greedy settings, the output distribution is stochastic. To strictly preserve $p(x)$, we adhere to the verified speculative sampling protocol. When a draft token $x_i$ is rejected, we sample a correction $x'_i$ from the residual distribution $p(x) - q(x)$ rather than deterministically selecting the top-$1$ token.
Crucially, unlike the greedy case, the sampled correction $x'_i$ may differ from the pre-retrieved target token $y_i$. To maintain distributional consistency, we strictly \textbf{discard the subsequent Target Guidance} ($y_{i+1}, \dots$) upon rejection, as these tokens were generated conditionally on a specific path that is no longer valid. However, the \textbf{Multi-token Pre-verify} mechanism remains active for the prefix preceding the rejection point. This ensures that the system benefits from early filtering of divergent tokens while strictly adhering to the target distribution.

\subsubsection{Empirical Verification}

To empirically validate that \textsc{Double} introduces no deviation from the target model's output, we conducted exact-match accuracy tests on the GSM8K dataset across three temperature settings ($T \in \{0, 0.5, 1.0\}$).

As presented in Table~\ref{tab:lossless_verification} and~\ref{tab:lossless_verification1}, \textsc{Double} achieves accuracy scores identical to the Vanilla autoregressive baseline (within numerical hardware precision limits for \textsc{bfloat16}) across all temperatures. For instance, on LLaMA-3.3-70B across GSM8K, the accuracy remains invariant at 0.93 ($T=0$), 0.92 ($T=0.5$), and 0.89 ($T=1.0$) for \textsc{Double}. These results confirm that \textsc{Double} delivers substantial speedups 3.67$\times$ without compromising generation quality or altering the theoretical output distribution.

\begin{table}[h]
\centering
\caption{\textbf{Verification of Lossless Acceleration.} We compare the Exact Match (EM) accuracy and Speedup of \textsc{Double} against the Vanilla baseline on GSM8K. The results confirm that \textsc{Double} maintains output consistency with the target model across varying temperatures. (The minor drops stem from the inherent uncertainty of LLMs and the precision limits of hardware numerics.)}
\label{tab:lossless_verification}
\vspace{5pt}
\resizebox{\linewidth}{!}{%
\begin{tabular}{l l cc cc cc}
\toprule
\multirow{2}{*}{\textbf{Models}} & \multirow{2}{*}{\textbf{Methods}} & \multicolumn{2}{c}{\textbf{Temperature=0}} & \multicolumn{2}{c}{\textbf{Temperature=0.5}} & \multicolumn{2}{c}{\textbf{Temperature=1.0}} \\
\cmidrule(lr){3-4} \cmidrule(lr){5-6} \cmidrule(lr){7-8} 
& & Acc. & Speedup & Acc. & Speedup & Acc. & Speedup \\
\midrule
\multirow{2}{*}{Qwen-32B} 
& Vanilla & 0.94 & 1.00$\times$ & 0.92 & 1.00$\times$ & 0.89 & 1.00$\times$ \\
& Sps & 0.94 & 1.00$\times$ & 0.91 & 1.00$\times$ & 0.88 & 1.00$\times$ \\
& \cellcolor{pink!30}\textbf{\textsc{Double}} & \cellcolor{pink!30}\textbf{0.94} & \cellcolor{pink!30}\textbf{2.30$\times$} & \cellcolor{pink!30}\textbf{0.91} & \cellcolor{pink!30}\textbf{2.12$\times$} & \cellcolor{pink!30}\textbf{0.89} & \cellcolor{pink!30}\textbf{2.03$\times$} \\
\midrule
\multirow{2}{*}{LLaMA-3.3-70B} 
& Vanilla & 0.93 & 1.00$\times$ & 0.92 & 1.00$\times$ & 0.90 & 1.00$\times$ \\
& Sps & 0.94 & 1.00$\times$ & 0.91 & 1.00$\times$ & 0.88 & 1.00$\times$ \\
& \cellcolor{pink!30}\textbf{\textsc{Double}} & \cellcolor{pink!30}\textbf{0.93} & \cellcolor{pink!30}\textbf{4.41$\times$} & \cellcolor{pink!30}\textbf{0.92} & \cellcolor{pink!30}\textbf{4.23$\times$} & \cellcolor{pink!30}\textbf{0.89} & \cellcolor{pink!30}\textbf{3.89$\times$} \\
\bottomrule
\end{tabular}
}
\vspace*{-0.1in}
\end{table}

\begin{table}[h]
\centering
\caption{\textbf{More results of Lossless Acceleration.} We compare the Exact Match (EM) accuracy and Speedup of \textsc{Double} against the Vanilla baseline on Deepseek1.3\&33B across three benchmarks. }
\label{tab:lossless_verification1}
\vspace{5pt}
\resizebox{\linewidth}{!}{%
\begin{tabular}{l l cc cc cc}
\toprule
\multirow{2}{*}{\textbf{Benchmarks}} & \multirow{2}{*}{\textbf{Methods}} & \multicolumn{2}{c}{\textbf{Temperature=0}} & \multicolumn{2}{c}{\textbf{Temperature=0.5}} & \multicolumn{2}{c}{\textbf{Temperature=1.0}} \\
\cmidrule(lr){3-4} \cmidrule(lr){5-6} \cmidrule(lr){7-8} 
& & Acc. & Speedup & Acc. & Speedup & Acc. & Speedup \\
\midrule
\multirow{2}{*}{HumanEval} 
& Vanilla & 0.65 & 1.00$\times$ & 0.62 & 1.00$\times$ & 0.59 & 1.00$\times$ \\
& Sps & 0.65& 2.15$\times$ & 0.63 & 1.95$\times$ & 0.58 & 1.81$\times$ \\
& \cellcolor{pink!30}\textbf{\textsc{Double}} & \cellcolor{pink!30}\textbf{0.65} & \cellcolor{pink!30}\textbf{4.45$\times$} & \cellcolor{pink!30}\textbf{0.63} & \cellcolor{pink!30}\textbf{4.22$\times$} & \cellcolor{pink!30}\textbf{0.59} & \cellcolor{pink!30}\textbf{4.13$\times$} \\
\midrule
\multirow{2}{*}{GSM8K} 
& Vanilla & 0.41 & 1.00$\times$ & 0.39 & 1.00$\times$ & 0.35 & 1.00$\times$ \\
& Sps & 0.41 & 1.82$\times$ & 0.39 & 1.70$\times$ & 0.35 & 1.64$\times$ \\
& \cellcolor{pink!30}\textbf{\textsc{Double}} & \cellcolor{pink!30}\textbf{0.41} & \cellcolor{pink!30}\textbf{3.83$\times$} & \cellcolor{pink!30}\textbf{0.39} & \cellcolor{pink!30}\textbf{3.62$\times$} & \cellcolor{pink!30}\textbf{0.35} & \cellcolor{pink!30}\textbf{3.43$\times$} \\
\midrule
\multirow{2}{*}{CNN/DM} 
& Vanilla & 0.37 & 1.00$\times$ & 0.32 & 1.00$\times$ & 0.30 & 1.00$\times$ \\
& Sps & 0.37 & 2.05$\times$ & 0.32 & 1.90$\times$ & 0.31 & 1.78$\times$ \\
& \cellcolor{pink!30}\textbf{\textsc{Double}} & \cellcolor{pink!30}\textbf{0.37} & \cellcolor{pink!30}\textbf{4.05$\times$} & \cellcolor{pink!30}\textbf{0.32} & \cellcolor{pink!30}\textbf{3.93$\times$} & \cellcolor{pink!30}\textbf{0.31} & \cellcolor{pink!30}\textbf{3.85$\times$} \\
\bottomrule
\end{tabular}
}
\vspace*{-0.1in}
\end{table}

\subsection{Hierarchical Datastore}
\label{sec:appendix_datastore}
To optimize the trade-off between retrieval coverage and memory efficiency, while effectively leveraging verification signals, we implement a hierarchical multi-layered datastore. The detailed procedure is formalized in Algorithm~\ref{alg:retrieval_cache}.
\paragraph{Data Decontamination and Fairness}
We explicitly address potential concerns regarding the initialization of our prior datastore using ShareGPT.
First, our setup ensures strictly fair comparisons: ShareGPT serves as the standard training corpus for the strongest baseline, EAGLE-3~\citep{li2025eagle}, and as the reference corpus for Token Recycling~\citep{token-recycle-luo-2024}. By utilizing the same data source for our retrieval prior (frontier knowledge), we ensure that \textsc{Double} does not benefit from superior data quality compared to baselines.
Second, ShareGPT acts as generic prior knowledge. We strictly treat it as a frozen warmup set distinct from the evaluation benchmarks (HumanEval, GSM8K, MT-Bench). Since the retrieval process relies on strictly matching $n$-gram contexts from the input prompt (which comes from the test set) to the datastore, and our datastore contains only generic chat data, the risk of "test set leakage" is structurally minimized. Thus, the performance gains are attributable to the architectural efficiency of our method rather than memorization.

\begin{algorithm}[t]
    \caption{Retrieval Cache Mechanism}\label{alg:retrieval_cache}
    \renewcommand{\algorithmicrequire}{\textbf{Input:}}
    \definecolor{desc}{RGB}{128,128,128}
    \definecolor{hit}{RGB}{34,139,34}
    
    \begin{algorithmic}[1]
        \REQUIRE{Input tokens $\mathbf{x}$, Max n-gram size $N$, Prediction length $K$}
        
        \STATE \textcolor{desc}{$\triangleright$ \textbf{Cache Structure}}
        \STATE $\mathcal{C}_{prefix}$ \quad \textcolor{desc}{$\triangleright$ Persistent prefix cache (warmup data)}
        \STATE $\mathcal{C}_{dynamic}$ \quad \textcolor{desc}{$\triangleright$ Dynamic cache (current task tokens)}
        \STATE $\mathcal{C}_{rejected}$ \quad \textcolor{desc}{$\triangleright$ Rejected tokens cache}
        
        \STATE \textcolor{desc}{$\triangleright$ \textbf{N-gram Retrieval (Priority Search)}}
        \FOR{$n = \min(N, |\mathbf{x}|)$ down to $1$}
            \STATE $\text{ngram} \leftarrow \mathbf{x}[-n:]$ \quad \textcolor{desc}{$\triangleright$ Extract suffix}
            
            \STATE \textcolor{desc}{$\triangleright$ Search in prefix cache first}
            \IF{ngram $\in \mathcal{C}_{prefix}$}
                \STATE \textcolor{hit}{$\checkmark$ Return next $K$ tokens from $\mathcal{C}_{prefix}$}
            \ENDIF
            
            \STATE \textcolor{desc}{$\triangleright$ Then search in dynamic cache}
            \IF{ngram $\in \mathcal{C}_{dynamic}$}
                \STATE \textcolor{hit}{$\checkmark$ Return next $K$ tokens from $\mathcal{C}_{dynamic}$}
            \ENDIF
            
            \STATE \textcolor{desc}{$\triangleright$ Finally search in rejected cache}
            \IF{ngram $\in \mathcal{C}_{rejected}$}
                \STATE \textcolor{hit}{$\checkmark$ Return next $K$ tokens from $\mathcal{C}_{rejected}$}
            \ENDIF
        \ENDFOR
        
        \STATE \textcolor{desc}{$\triangleright$ Fallback: search in input sequence itself}
        \RETURN Original PLD search in $\mathbf{x}$
        
        \STATE \textcolor{desc}{$\triangleright$ \textbf{Cache Update}}
        \STATE $\mathcal{C}_{dynamic} \leftarrow \mathcal{C}_{dynamic} \cup \{\text{accepted tokens}\}$
        \STATE $\mathcal{C}_{rejected} \leftarrow \mathcal{C}_{rejected} \cup \{\text{rejected tokens}\}$
    \end{algorithmic}
\end{algorithm}

\paragraph{Clarification on Memory Overhead}
Concerns regarding retrieval-based methods often stem from the presumption of unbounded datastore growth. We emphasize that \textsc{Double} employs a strictly \textbf{Lightweight \& Ephemeral} storage protocol designed with minimal resource footprint:
\begin{itemize}[leftmargin=*, itemsep=0pt, topsep=2pt]
    \item \textbf{Static Prior (Initialization):} As detailed in the methodology, the prior knowledge is serialized as a compact local file (approx. 9.5 MB). It is loaded once into CPU RAM as a read-only trie structure, incurring negligible system overhead.
    \item \textbf{Ephemeral Dynamic Datastore:} distinct from RAG frameworks that maintain persistent, high-dimensional vector databases, our dynamic datastore records only integer Token IDs rather than hidden states. Crucially, the dynamic layers ($\mathcal{C}_{\text{dynamic}}$ and $\mathcal{C}_{\text{rejected}}$) are session-scoped: they are initialized at the start of a generation request and immediately \emph{flushed} upon completion. Even during long-context generation (e.g., 8k tokens), this integer-based mapping consumes only a few megabytes of CPU RAM, effectively circumventing VRAM scarcity issues on consumer-grade hardware.
\end{itemize}

\begin{table}[h]
\centering
\caption{\textbf{Ablation Study on Rejected Token Cache.} Performance comparison on Qwen3-32B (MT-Bench). Incorporating the Rejected Cache significantly improves the retrieval hit rate, leading to superior speedups.}
\label{tab:ablation_rejected}
\resizebox{\columnwidth}{!}{%
\begin{tabular}{lccccccc}
\toprule
\textbf{Configuration} & \textbf{Writing} & \textbf{Roleplay} & \textbf{Reasoning} & \textbf{Math} & \textbf{Coding} & \textbf{Extraction} & \textbf{Avg.} \\
\midrule
\textsc{Double} (w/o Rejected Cache) & 2.15$\times$ & 2.10$\times$ & 2.38$\times$ & 2.05$\times$ & 2.18$\times$ & 1.95$\times$ & 2.14$\times$ \\
\midrule
\cellcolor{pink!30}\textbf{\textsc{Double} (Full)} & \cellcolor{pink!30}\textbf{2.37$\times$} & \cellcolor{pink!30}\textbf{2.23$\times$} & \cellcolor{pink!30}\textbf{2.55$\times$} & \cellcolor{pink!30}\textbf{2.21$\times$} & \cellcolor{pink!30}\textbf{2.33$\times$} & \cellcolor{pink!30}\textbf{2.03$\times$} & \cellcolor{pink!30}\textbf{2.30$\times$} \\
\bottomrule
\end{tabular}
}
\vspace*{-0.1in}
\end{table}

\begin{figure}[t]
    \centering
    \includegraphics[width=0.48\textwidth]{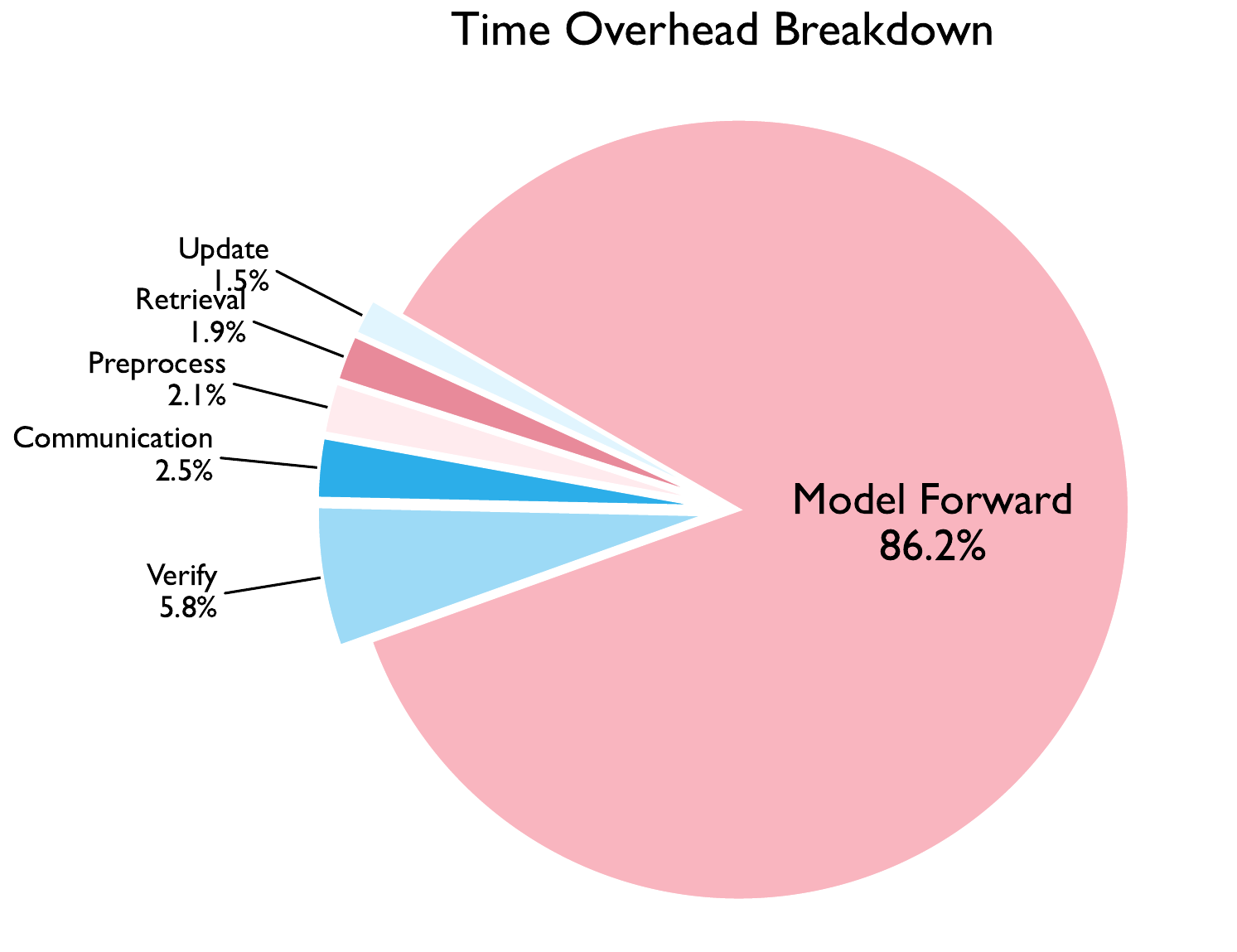}
\vspace{-0.25in}
\caption{Profiling results showing that \textsc{Double} incurs minimal overhead: retrieval (1.9\%) and communication (2.5\%) remain negligible compared to model forward (86.2\%) in single \textbf{Retrieval Forward}.}
    \label{fig:time_overhead}
    \vspace{-0.2in}
\end{figure}

\paragraph{Analysis of Rejected Tokens}
A legitimate concern is that caching rejected tokens might pollute the retrieval pool with suboptimal candidates. However, within the context of speculative decoding, a rejection primarily signifies a divergence from the target model's current output rather than a semantic error. These tokens often represent valid synonymic variations or near misses. By caching these rejected yet high-probability tokens, \textsc{Double} preserves valid alternative generation paths. This allows the model to recall and reuse them when similar contexts recur, effectively transforming 'waste' into valuable candidates and boosting the retrieval hit rate.

We empirically validate this design in Table~\ref{tab:ablation_rejected}. The inclusion of the Rejected Token Cache ($\mathcal{C}_{\text{rejected}}$) yields consistent performance gains across all benchmarks (Avg. $2.30\times$ vs. $2.14\times$). This confirms that utilizing the rejected history provides valuable alternative guidance that augments the retrieval process.

\subsection{Time Consumption}
\label{sec:appendix_time}
To assess the runtime overhead of our proposed method, we profile the latency breakdown of a single \textbf{Retrieval Forward}, as illustrated in Figure~\ref{fig:time_overhead}. The results indicate that the inference process is dominated by the \textbf{Model Forward} phase (86.2\%), confirming that the backbone model's computation remains the primary resource consumer. Crucially, the \textbf{Retrieval} mechanism introduces a mere \textbf{1.9\%} overhead. This negligible cost validates our design choice of using a lightweight, local datastore, ensuring that the retrieval-augmented drafting process does not become a bottleneck. 
However, we observe that auxiliary operations such as \textbf{Communication} (2.5\%) and KV-cache updates (Verify 5.8\%, Update 1.5\%) still constitute a noticeable portion of the latency. These overheads are largely attributed to the architectural constraints of the Python-based \texttt{transformers} library, which is suboptimal for inter-process communication and efficient memory management (e.g., KV-cache rollback). We anticipate that migrating \textsc{Double} to high-performance inference frameworks with custom CUDA kernels (e.g., vLLM or SGLang) will significantly mitigate these engineering overheads, pushing the actual speedup closer to the theoretical ceiling. We reserve this system-level optimization for future work to facilitate broader industrial deployment.

\subsection{Precision-Efficiency Dilemma}
\label{sec:appendix_dilemma}
\begin{figure}[t]
    \centering
    \includegraphics[width=0.48\textwidth]{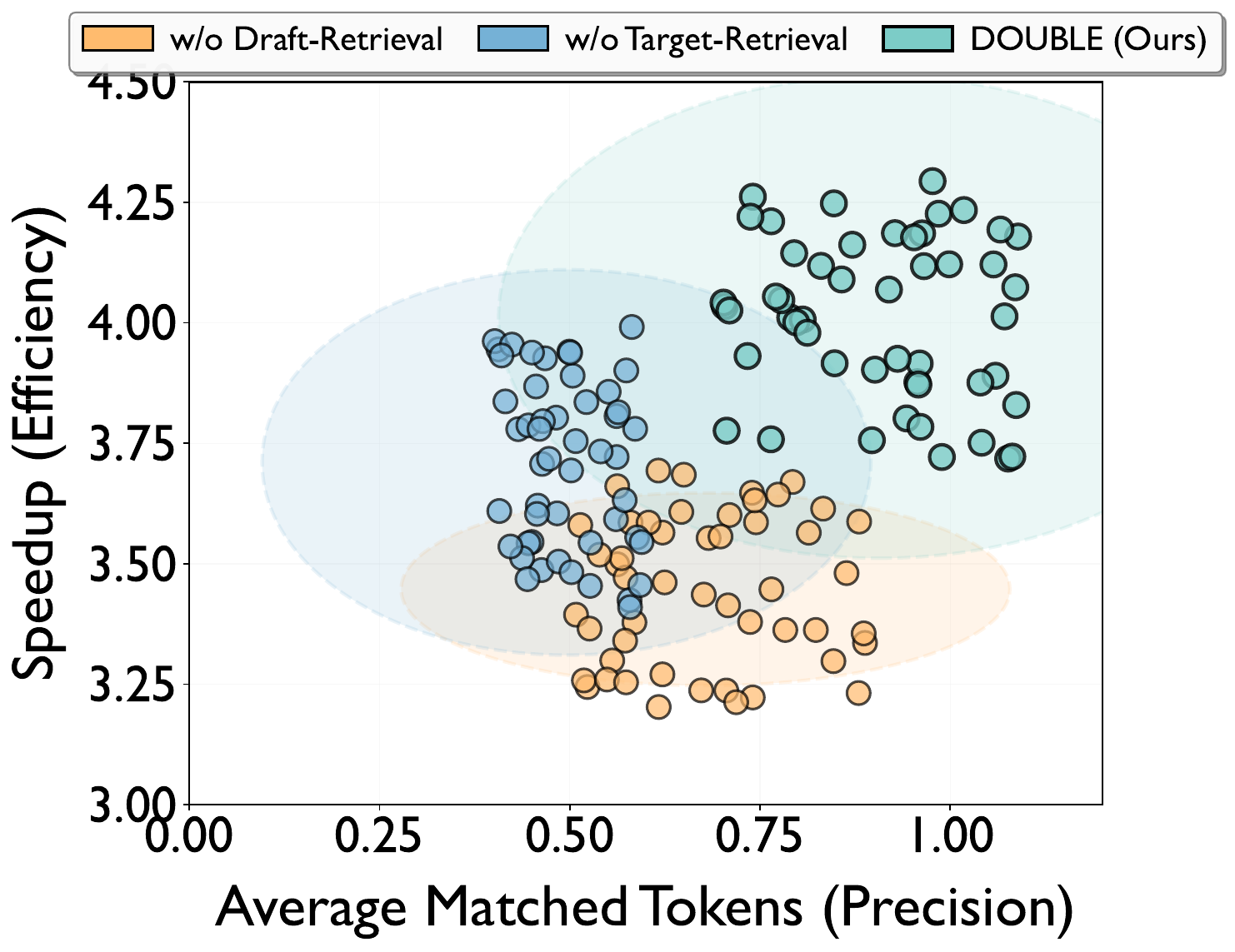}
    \vspace{-0.1in}
    \caption{\textbf{Precision-Efficiency Analysis.} The scatter plot compares \textsc{Double} against single-sided ablations. \textsc{Double} (Teal) resolves the dilemma by achieving both high Precision (AMT) and Efficiency (Speedup). Note that architectural overheads in \texttt{transformers} (e.g., KV rollback) create a slight gap between AMT gains and wall-time speedup, which we aim to address in future high-performance implementations.}
    \label{fig:dilemma}
    \vspace{-0.2in}
\end{figure}
To illustrate how \textsc{Double} navigates the trade-off between retrieval quality and inference speed, we visualize the relationship between Average Matched Tokens (AMT), a proxy for Precision and Wall-time Speedup (Efficiency) in Figure~\ref{fig:dilemma}. The scatter plot, derived from LLaMA-3.3-70B inference traces, delineates three distinct performance regimes:

\begin{itemize}[leftmargin=*, itemsep=0pt, topsep=2pt]
    \item \textbf{Low Efficiency Regime (\textit{w/o Draft-Retrieval}):} Marked in \textcolor{orange}{orange}, this variant depends on autoregressive drafting. While it maintains respectable precision, its efficiency is strictly constrained by the theoretical speed ceiling $C$, clustering in the lower-speed region.
    \item \textbf{Low Precision Regime (\textit{w/o Target-Retrieval}):} Marked in \textcolor{cyan}{blue}, this variant utilizes draft-side retrieval but lacks authoritative target guidance. Constrained by the limited capacity of the lightweight draft model, it suffers from frequent verification failures, resulting in low AMT and suboptimal speedups.
    \item \textbf{Optimal Regime (\textsc{Double}):} Marked in \textcolor{teal}{teal}, our method occupies the top-right quadrant, representing the Pareto-optimal frontier. By synergizing the high throughput of draft retrieval with the precise guidance of the target model, \textsc{Double} simultaneously maximizes AMT ($>0.9$) and Speedup ($>4.0\times$), effectively resolving the dilemma.
\end{itemize}

\paragraph{Engineering Gap Analysis}
Notably, we observe a divergence between the high AMT values and the realized wall-time speedups across all retrieval-based methods. This discrepancy stems from the architectural limitations of the standard \texttt{transformers} library. Operations such as inter-process communication and, more critically, the costly KV-cache rollback mechanisms upon rejection, incur non-trivial latency overheads. These engineering bottlenecks partially offset the throughput gains derived from high retrieval precision. 
To bridge this gap, future work will focus on migrating \textsc{Double} to high-performance inference engines like vLLM or SGLang. By leveraging custom CUDA kernels and optimized memory management, we aim to eliminate these overheads and fully translate algorithmic precision into industrial-grade acceleration.

\subsection{Analysis of Memory-Constrained Scenarios}
\label{sec:appendix_memory}

This section evaluates the adaptability of \textsc{Double} across varying resource environments, distinguishing it from traditional sequential methods. Unlike ``draft-then-verify'' paradigms that often underutilize hardware, \textsc{Double} is designed to exploit parallel computation. We categorize deployment scenarios based on memory availability:

\paragraph{Resource-Abundant Scenarios}
\textsc{Double} is optimal for multi-GPU cloud environments or cloud-edge collaborative setups where computational resources are ample. In these settings, the draft and target models are deployed on independent processors (e.g., separate GPUs or heterogeneous CPU/GPU clusters). This physical isolation eliminates memory contention, preventing the draft model from slowing down the target model's primary inference loop. Our main experiments (Section~\ref{sec:main_results}) reflect this ideal configuration. Future integrations with Tensor Parallelism (TP)~\citep{pd_2024} could further amplify acceleration in these robust environments.

\paragraph{Resource-Constrained Scenarios (Shared GPUs)}
In scenarios where the draft and target models must share GPUs (e.g., a 33B model spanning two 40GB A100s, with a 1.3B draft model co-located), resource contention becomes a bottleneck. To mitigate this, we adopt a modified Pipeline Parallelism (PP) strategy inspired by PEARL~\citep{liu2024parallel}. 
By interleaving computation, the draft model generates tokens on the idle GPU while the target model executes on the active GPU. For instance, while the target model computes on GPU 0, the draft model operates in parallel on GPU 1; they efficiently swap roles as the target model's pipeline progresses. 
Although this introduces minor communication overhead, it effectively circumvents memory contention. As shown in Table~\ref{tab:memorycons}, experiments with Deepseek 1.3B \& 33B on Spec-Bench demonstrate that \textsc{Double} with PP retains \textbf{89.76\%} of its peak performance, significantly outperforming vanilla SD (SpS) even under constrained settings.

\begin{table}[h]
\centering
\caption{Performance retention of \textsc{Double} under memory-constrained scenarios (Deepseek 1.3B \& 33B on MT-Bench). Using Pipeline Parallelism (PP), \textsc{Double} maintains $\sim$90\% of its ideal acceleration.}
\label{tab:memorycons}
\resizebox{\columnwidth}{!}{%
\begin{tabular}{lccccccc}
\toprule
\textbf{Methods} & \textbf{Writing} & \textbf{Roleplay} & \textbf{Reasoning} & \textbf{Math} & \textbf{Coding} & \textbf{Extraction} & \textbf{Avg.} \\
\midrule
SpS (Vanilla SD) & 1.89$\times$ & 2.08$\times$ & 2.14$\times$ & 1.97$\times$ & 2.23$\times$ & 1.75$\times$ & 2.01$\times$ \\
\midrule
\cellcolor{pink!30}\textbf{\textsc{Double} (Ideal)} & \cellcolor{pink!30}\textbf{3.86$\times$} & \cellcolor{pink!30}\textbf{4.25$\times$} & \cellcolor{pink!30}\textbf{4.38$\times$} & \cellcolor{pink!30}\textbf{4.01$\times$} & \cellcolor{pink!30}\textbf{4.56$\times$} & \cellcolor{pink!30}\textbf{3.54$\times$} & \cellcolor{pink!30}\textbf{4.10$\times$} \\
\cellcolor{pink!30}\textbf{\textsc{Double} (w/ PP)} & \cellcolor{pink!30}\textbf{3.48$\times$} & \cellcolor{pink!30}\textbf{3.74$\times$} & \cellcolor{pink!30}\textbf{3.94$\times$} & \cellcolor{pink!30}\textbf{3.69$\times$} & \cellcolor{pink!30}\textbf{3.96$\times$} & \cellcolor{pink!30}\textbf{3.19$\times$} & \cellcolor{pink!30}\textbf{3.67$\times$} \\
\midrule
\cellcolor{blue!15}{\textbf{Retention}} & \cellcolor{blue!15}{\textbf{90.2\%}} & \cellcolor{blue!15}{\textbf{88.0\%}} & \cellcolor{blue!15}{\textbf{90.0\%}} & \cellcolor{blue!15}{\textbf{92.0\%}} & \cellcolor{blue!15}{\textbf{86.8\%}} & \cellcolor{blue!15}{\textbf{90.1\%}} & \cellcolor{blue!15}{\textbf{89.5\%}} \\
\bottomrule
\end{tabular}
}
\vspace*{-0.1in}
\end{table}

\paragraph{Single-GPU Scenarios (Extreme Constraint)}
In single-GPU environments, \textsc{Double} gracefully degrades to a serialized workflow, executing retrieval-augmented drafting followed by target verification. Table~\ref{tab:singlegpu} compares this mode against PEARL (which reverts to Vanilla SD) on Deepseek 1.3B \& 33B. Even without parallel execution, \textsc{Double}'s retrieval mechanism ensures superior drafting precision, outperforming the baseline by a clear margin ($1.81\times$ vs. $1.64\times$). Furthermore, advanced frameworks like Nano-vLLM offer a pathway to restore flexibility: by leveraging Gloo for intra-device process orchestration and utilizing concurrent CUDA streams, \textsc{Double} can achieve logical parallelism for verification even on a single device, maximizing hardware utilization.

\begin{table}[h]
\centering
\caption{Performance in Single-GPU scenarios (Deepseek 1.3B \& 33B on MT-Bench). \textsc{Double} outperforms PEARL (Vanilla SD) even without parallel execution capabilities.}
\label{tab:singlegpu}
\resizebox{\columnwidth}{!}{%
\begin{tabular}{lccccccc}
\toprule
\textbf{Methods} & \textbf{Writing} & \textbf{Roleplay} & \textbf{Reasoning} & \textbf{Math} & \textbf{Coding} & \textbf{Extraction} & \textbf{Avg.} \\
\midrule
PEARL (SpS) & 1.89$\times$ & 2.08$\times$ & 2.14$\times$ & 1.97$\times$ & 2.23$\times$ & 1.75$\times$ & 2.01$\times$ \\
\midrule
\cellcolor{pink!30}\textbf{\textsc{Double}} & \cellcolor{pink!30}\textbf{2.37$\times$} & \cellcolor{pink!30}\textbf{2.23$\times$} & \cellcolor{pink!30}\textbf{2.55$\times$} & \cellcolor{pink!30}\textbf{2.21$\times$} & \cellcolor{pink!30}\textbf{2.33$\times$} & \cellcolor{pink!30}\textbf{2.03$\times$} & \cellcolor{pink!30}\textbf{2.29$\times$} \\
\bottomrule
\end{tabular}
}
\vspace*{-0.15in}
\end{table}

\section{Detailed Related Works}
While Large Language Models (LLMs) have achieved remarkable success in various benchmarks~\citep{team2023gemini,gpt3-brown-2020,qwen2-yang-2024,guo2025deepseek,MachineLearningI}, their deployment is severely constrained by their auto-regressive token-by-token generation.
\vspace*{-0.05in}
\paragraph{Efficient LLM Architectures.}
Substantial research has been directed toward accelerating inference through structural model optimizations. Key approaches include:
1) \textbf{Model Distillation}~\citep{minitron-sreenivas-2024, compact-llm-muralidharan-2024,lu2025onpolicydistillation,sanh2019distilbert,hinton2015distilling}, which compresses knowledge from a large teacher model into a compact student model to boost speed while maintaining capability;
2) \textbf{Quantization}~\citep{frantar2023sparsegpt, smoothquant-xiao-2023, awq-lin-2024, spinquant-liu-2024, quarot-ashkboos-2024}, which reduces parameter precision to lower storage footprint and minimizes data transfer latency from HBM to on-chip memory; and
3) \textbf{Pruning}~\citep{frantar2023sparsegpt, dao2022flashattention, shortgpt-men-2024, llm-streamline-chen-2024, hu2022lora, wanda-sun-2024, RIA-zhang-2024}, which eliminates redundant parameters. Specifically, structured pruning is often coupled with distillation to train efficient lightweight models to reduce memory access overhead.
Despite these advances in reducing computational complexity, LLM inference remains inherently \textit{memory-bound} due to the sequential nature of auto-regressive decoding. Furthermore, these structural modifications often necessitate extensive retraining or specialized hardware support, and few strategies effectively resolve the memory bandwidth bottleneck without compromising performance.

\label{appendix:related_work}
\vspace*{-0.05in}
\paragraph{Speculative Decoding}
While Speculative Decoding (SD)~\citep{agrawal2024adaedl,chen2024sequoia,huang2024specdec++,liu2024kangaroo,spec-bench-xia-2024,xia2024swift,yang2025longspec,huang2025jakiro,liu2026talon}, has demonstrated significant acceleration to alleviate \textit{memory-bound} with lossless generalization, maximizing the draft token acceptance rate remains a critical challenge. Existing literature primarily addresses this through training-based or training-free alignment strategies. For instance, Medusa~\citep{cai2024medusa} introduces auxiliary decoding heads, whereas EAGLE~\citep{li2024eagle} and Glide~\citep{du2024glide} leverage target model features to enhance drafting precision. Similarly, SpecInfer~\citep{chen2023accelerating} employs tree-based attention to verify multiple candidates. On the training-free front, methods like Draft\&Verify~\citep{zhang2024draft} utilize a subset of target model layers as a proxy for the draft model; however, the limited number of skipped layers constrains the overall speedup. More recently, EAGLE3~\citep{li2025eagle} has redefined the scaling laws for small model training via test-time training and feature extrapolation, gaining widespread industrial adoption. Nevertheless, a fundamental limitation persists across these approaches: their adherence to a sequential \emph{draft-then-verify} paradigm imposes an inherent mutual waiting bottleneck.
\vspace*{-0.05in}
\paragraph{Parallel Speculative Decoding} 
To mitigate the hardware underutilization in serialized execution, recent works have shifted toward parallel paradigms. PEARL~\citep{liu2024parallel} introduces a pipelined framework that enables the draft model to generate subsequent tokens concurrently with the target model's verification of the initial draft. Building on this, SpecBranch~\citep{shen2025speculative} incorporates fine-grained dynamic length adjustments and robust multi-branch fallback strategies to further optimize parallel efficiency. In the distributed landscape, DSI~\citep{timor2024distributed} proposes a parallel framework that orchestrates the temporal overlap between target and drafter instances. However, the efficacy of these approaches remains constrained by the theoretical speedup limits of specific model pairings and the severe penalties associated with rollback, which inevitably disrupt the pipeline.
\paragraph{Retrieval Speculative Decoding} Retrieval-based methods, which generate drafts via $n$-gram matching, have emerged as a distinct branch of SD. These methods generally bifurcate into two categories: 1) Target-Side Self-SD, where approaches like Lookahead Decoding~\citep{fu2024break}, PLD~\citep{pld-saxena-2023}, REST~\citep{rest-he-2024}, and LogitSpec~\citep{liu2025logitspec} focus on accelerating the target model intrinsically. 2) Draft-Side Acceleration, where Ouroboros~\citep{zhao2024ouroboros} stands as the sole work employing lookahead logic to expedite draft model generation.  However, existing retrieval mechanisms operate in isolation on either the target or draft model, creating an inherent \emph{Retrieval Precision-Efficiency Dilemma}. To address these challenges, we propose \textbf{\textsc{Double}}, which resolves this dilemma by orchestrating a synergistic dual-retrieval strategy. By leveraging the complementary strengths of both models simultaneously, \textsc{Double} breaks the theoretical speedup ceiling.
\vspace*{-0.05in}
\paragraph{More Discussions about EAGLE} 
We acknowledge that EAGLE-3's dedicated draft model is a significant but orthogonal design choice for PSD. \textsc{Double} currently utilizes a standard draft model; however, this opens a promising avenue for future work to integrate on-policy online distillation or EAGLE's draft design, developing specialized draft models to further enhance the parallel framework.
\vspace*{-0.05in}
\paragraph{Discussions on Future Work.} 
Although speculative decoding has achieved remarkable success in accelerating pure-text generation, its potential across diverse downstream applications remains to be fully explored. Looking forward, we envision that Parallel Speculative Decoding (PSD) paradigms, exemplified by \textsc{Double}, can be seamlessly adapted into several promising avenues. \textbf{1) multimodal LLM acceleration}, where PSD can be tailored to efficiently serve large vision-language and video models through visual-semantic guidance, KV cache compression, and novel parallel decoding architectures~\cite{zhang2026efficientinferencelargevisionlanguage, kong2026parallelvlm, kong2026vision, ji2026foresttreeslooselyspeculative, zeng2026hybridkvhybridkvcache, wang2025fakesvvlmtamingvlmdetecting,wang2026streammecolongtermagentmemory}. \textbf{2) complex agentic reasoning}, where accelerated on-the-fly logical deduction, dynamic policy exploration, and multi-perspective verification are critical for agent workflows~\cite{sun2025causalabstain,sun2026factecausalityinspiredevaluationtrustworthy,wu2024beyond, zhang2026logical, zhang2025ambiguity, wu2026atlas, wu2026ssl, wu2026spark, wang2026perm}. \textbf{3) broader efficiency optimizations}, including synergies with native parallel reading mechanisms inside Transformers and pipelined multi-DNN execution on heterogeneous hardware, to push the limits of modern inference systems~\cite{wang2026fbs, shen2025hetero, an2026flowhijackdynamicsawarebackdoorattack,shen2025batch,shen2025flowmesh}.

\end{document}